\documentclass[journal]{IEEEtran}
\usepackage{amsmath,amsthm,amsfonts,mathrsfs}
\usepackage{algorithmic}
\usepackage{array}
\usepackage[caption=false,font=normalsize,labelfont=sf,textfont=sf]{subfig}
\usepackage[pagebackref=true,breaklinks=true,colorlinks,bookmarks=false,citecolor=blue]{hyperref}
\usepackage{eurosym,fancyhdr,CJK,multicol,graphics,indentfirst,color,bm,upgreek,booktabs,graphicx}
\usepackage{textcomp}
\usepackage{stfloats}
\usepackage{url}
\usepackage{verbatim}
\usepackage{graphicx}
\usepackage{cite}
\usepackage{color}
\usepackage{multirow}
\usepackage{multicol}
\usepackage{makecell}
\usepackage{booktabs}
\usepackage{arydshln}
\usepackage{bbding}
\usepackage{amssymb}
\usepackage{pifont}
\usepackage{bigstrut}
\usepackage{rotating}
\usepackage[table]{xcolor}
\usepackage{colortbl} 

\newcommand{\ie}{\textit{i}.\textit{e}.}
\newcommand{\eg}{\textit{e}.\textit{g}.}
\newcommand{\etal}{\textit{et al}.}
\newcommand{\cmark}{\ding{51}}
\newcommand{\xmark}{\ding{55}}

\definecolor{tabletitle}{gray}{.8}
\definecolor{mygray}{gray}{.92}
\definecolor{ours}{gray}{.95}
\definecolor{mygreen}{RGB}{0,136,51}
\definecolor{myblue}{RGB}{0,102,204}
\definecolor{myred}{RGB}{202,0,0}

\ifCLASSINFOpdf
\else
\fi
\begin{document}
%
\title{Improving Misaligned Multi-modality Image Fusion with One-stage Progressive Dense Registration}
%
%
%

\author{Di Wang, Jinyuan Liu,~\IEEEmembership{Member,~IEEE}, Long Ma,~\IEEEmembership{Member,~IEEE}, Risheng Liu,~\IEEEmembership{Member,~IEEE}, and Xin Fan,~\IEEEmembership{Senior Member,~IEEE}
\thanks{This work is partially supported by the National Natural Science Foundation of China (Nos. U22B2052 and 62027826) and the Fundamental Research Funds for the Central Universities. (Corresponding author: X. Fan).}
\thanks{D. Wang and L. Ma are with the School of Software Technology, Dalian University of Technology, Dalian, 116024, China. (e-mail: diwang1211@mail.dlut.edu.cn; malone94319@gmail.com).}
\thanks{J. Liu is with the School of Mechanical Engineering, Dalian University of Technology, Dalian, 116024, China. (e-mail: atlantis918@hotmail.com).}
\thanks{R. Liu and X. Fan are with the DUT-RU International School of Information Science \& Engineering, Dalian University of Technology, Dalian, 116024, China.
		(e-mail: rsliu@dlut.edu.cn; xin.fan@dlut.edu.cn).}
  }

%
%

\markboth{Journal of \LaTeX\ Class Files,~Vol.~14, No.~8, August~2015}%
{Shell \MakeLowercase{\textit{et al.}}: Bare Demo of IEEEtran.cls for IEEE Journals}
%



\maketitle

\begin{abstract}
Misalignments between multi-modality images pose challenges in image fusion, manifesting as structural distortions and edge ghosts.
Existing efforts commonly resort to registering first and fusing later, typically employing two cascaded stages for registration,~\ie, coarse registration and fine registration. Both stages directly estimate the respective target deformation fields.
In this paper, we argue that the separated two-stage registration is not compact, and the direct estimation of the target deformation fields is not accurate enough.
To address these challenges, we propose a Cross-modality Multi-scale Progressive Dense Registration (C-MPDR) scheme, which accomplishes the coarse-to-fine registration exclusively using a one-stage optimization, thus improving the fusion performance of misaligned multi-modality images.
Specifically, two pivotal components are involved, a dense Deformation Field Fusion (DFF) module and a Progressive Feature Fine (PFF) module. The DFF aggregates the predicted multi-scale deformation sub-fields at the current scale, while the PFF progressively refines the remaining misaligned features. Both work together to accurately estimate the final deformation fields.
In addition, we develop a Transformer-Conv-based Fusion (TCF) subnetwork that considers local and long-range feature dependencies, allowing us to capture more informative features from the registered infrared and visible images for the generation of high-quality fused images.
%
Extensive experimental analysis demonstrates the superiority of the proposed method in the fusion of misaligned cross-modality images.	 
The code will be available at~\url{https://github.com/wdhudiekou/IMF}.
\end{abstract}


%
\IEEEpeerreviewmaketitle

\begin{figure}[t]
	\footnotesize
	\begin{center}
		\begin{tabular}{ccc}
			\includegraphics[width = 0.31\linewidth]{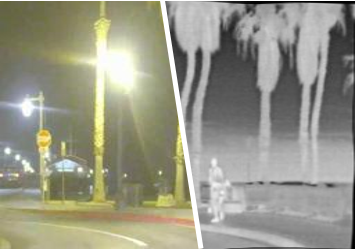} &
			\hspace{-4.5mm}
			\includegraphics[width = 0.31\linewidth]{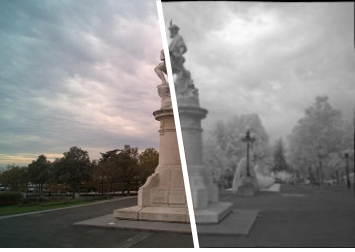} &
			\hspace{-4.5mm}
			\includegraphics[width = 0.31\linewidth]{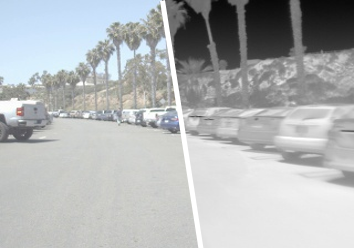}\\
			
			(a) Synthetic VIS/IR &
			\hspace{-4.5mm} (b) Synthetic VIS/NIR &
			\hspace{-4.5mm} (c) Real-world VIS/IR \\
			
			\includegraphics[width = 0.31\linewidth]{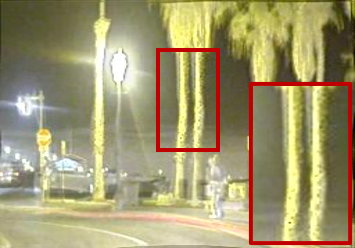} &
			\hspace{-4.5mm}
			\includegraphics[width = 0.31\linewidth]{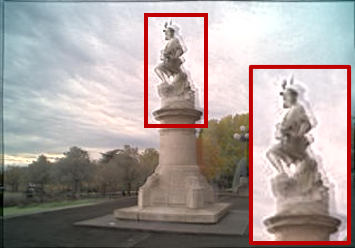} &
			\hspace{-4.5mm}
			\includegraphics[width = 0.31\linewidth]{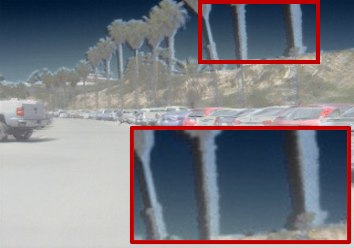}\\
			
			(d) CDDFuse~\cite{CDDFuse23} &
			\hspace{-4.5mm} (e) CDDFuse~\cite{CDDFuse23} &
			\hspace{-4.5mm} (f) CDDFuse~\cite{CDDFuse23} \\
			
			\includegraphics[width = 0.31\linewidth]{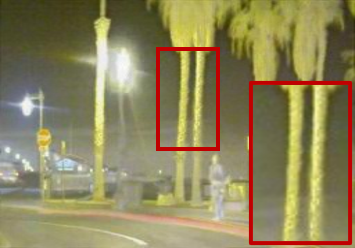} &
			\hspace{-4.5mm}
			\includegraphics[width = 0.31\linewidth]{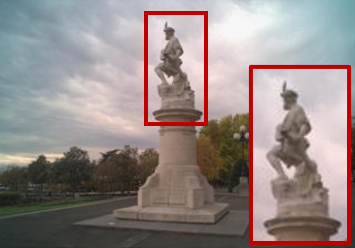} &
			\hspace{-4.5mm}
			\includegraphics[width = 0.31\linewidth]{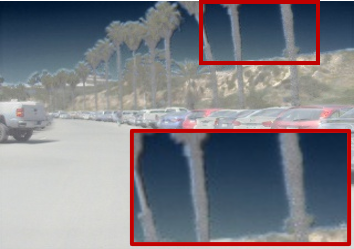}\\
			
			(g) IMF (Ours) &
			\hspace{-4.5mm} (h) IMF (Ours) &
			\hspace{-4.5mm} (i) IMF (Ours) \\
			
		\end{tabular}
	\end{center}
	\vspace{-2mm}
	\caption{Comparision of fusion results on different misaligned multi-modality images (\ie, synthetic misaligned NIR/IR-VIS images shown in (a) and (b), real-world misaligned IR-VIS images shown in (c)). CDDFuse~\cite{CDDFuse23} is a state-of-the-art image fusion method. By comparison, our IMF yields highly quality fusion images without structural distortions or edge ghosts.}
	\label{fig:first_img}
	\vspace{-2mm}	
\end{figure}

\section{Introduction}

Recent years have witnessed a growing interest in infrared and visible image fusion (IVIF), which aims to merge complementary information from infrared and visible sensors, thereby generating an image that thoroughly characterizes the scenario. 
Infrared sensor imaging utilizes thermal radiation emitted by objects and is unaffected by illumination variations, capable of compensating for the limitations of visible sensors in depicting critical attributes of scenes and objects, especially in adverse lighting conditions. 
Therefore, fusing them yields versatile and meaningful representations, which benefit practical applications ranging from object detection~\cite{TarDAL_22}, semantic segmentation, scene understanding~\cite{IRFS_23}, to autonomous driving.

Existing IVIF techniques~\cite{FGAN, DIDFuse_2020, PMGI, DDcGAN, RFN, MFEIF, U2Fusion} perform well in highlighting objects and preserving textures, yet they encounter difficulties in addressing structural distortions and edge ghosts in the fused image, stemmed from the misalignment between source images. 
As shown in Fig.~\ref{fig:first_img}(d)-(f), even CDDFuse~\cite{CDDFuse23}, one latest image fusion method, struggles with this dilemma.
The intrinsic reason is that these fusion methods are designed for manually pre-registered infrared-visible images and cannot accommodate even slight displacements.
A common solution is to ``\textit{register first, fuse later}". 
For image registration, the predominant methods~\cite{flownet,voxelmorph,dgcnet_19,glunet_20} rely on optical flow estimation, with the goal of estimating dense pixel correspondences between the source and target images.
Although increasingly refreshing registration accuracy, these methods suffer from a cliff-like drop in performance when directly applied to multi-modality images. This is attributed to the insurmountable modality discrepancies between source images.

To this end, three main lines have emerged:
\textit{i)} a modality promotion framework called CrossRAFT~\cite{crossRAFT_22} is proposed to extend the off-the-shelf mono-modality optical flow estimation models for multi-modality images. However, utilizing data augmentation to create multi-modality images limits its effectiveness in real-world multi-modality datasets.
\textit{ii)} a supervised versatile multi-modality image registration and fusion framework called SuperFusion~\cite{SuperFusion_22} is proposed. Despite its ability to effectively rectify geometric distortions of source images and generate fused images with semantic awareness, this study relies on strict supervision of GT deformation fields, which is unavailable for most public multi-modality datasets.
\textit{iii)} a unsupervised multi-modality image registration paradigm is favored by Nemar~\cite{NeMAR}, RFNet~\cite{RFNet_22}, and UMF~\cite{UMF}. Each of them employs image-to-image translation to implement style transfer between infrared and visible images, effectively reducing modality discrepancies.
%
Nevertheless, Nemar~\cite{NeMAR} trains image translation and registration simultaneously, leading to a complex and mutually destructive optimization.
Both RFNet~\cite{RFNet_22} and UMF~\cite{UMF} treat the reduction of modality discrepancies as an independent part to ensure that its function is not destroyed.
Specifically, RFNet encourages the cross-mapping of infrared and visible images to the domain of each other. UMF introduces a cross-modality perceptual style transfer (CPST) scheme to translate visible images into pseudo-infrared images. Both greatly
reduce the modality discrepancy, but they still have their own limitations.
For instance, RFNet~\cite{RFNet_22} performs coarse-to-fine registration through two stages,~\ie, coarse registration and fine registration, which are independent of each other. In contrast, the coarse-to-fine image registration method CGRP, proposed by UMF~\cite{UMF}, adopts a one-stage training manner. However, it solely relies on features extracted from two scales to estimate the final deformation fields, lacking a representation of global features. This renders it challenging to address even moderately larger deformations.
It is worth noting that both RFNet and UMF offer valuable insights. the CPST introduced by UMF intuitively reduces the cross-modality discrepancy from the pixel level. RFNet proves the existence of a mutually reinforced relationship between fine registration and fusion.

Motivated by this, this work aims to address the issue of misaligned multi-modality image fusion. Specifically, two key aspects pertaining to this issue are:
\textit{i) how to effectively register misaligned infrared and visible images in a one-stage coarse-to-fine manner.}
\textit{ii) How to fuse informative features from the registered infrared and visible images to generate a high-quality fused image, thereby reversely facilitating the registration performance.}
Therefore, we introduce an \textbf{I}mproving \textbf{M}isaligned infrared and visible image \textbf{F}usion framework (\textbf{IMF}) to effectively mitigate structural distortions and edge ghosts.
To be specific, we propose a Cross-modality Multi-scale Progressive Dense Registration (C-MPDR) scheme to register misaligned infrared and visible images. In the C-MPDR, we first translate the given visible image into a pseudo-infrared image by the CPST in UMF~\cite{UMF}, effectively reducing the large modality discrepancies. Subsequently, we utilize the MPDR subnetwork to register the source-distorted infrared image and the pseudo-infrared image. Here, a two-stream multi-scale feature extractor is incorporated, wherein the top-level features have a small size and a global receptive field. By performing coarse registration on top-level features, severe global pixel shifts can be effectively mitigated. Shallow features serve to model local attributes, and fine registration on them becomes instrumental in the correction of subtle local deformations. Therefore, it is expected to achieve the coarse-to-fine cross-modality image registration in a one-stage optimization.
To ensure a more accurate estimation of the target deformation fields, we innovatively develop coarse-to-fine learning from both deformation fields and feature perspectives.
We propose two modules, namely the dense Deformation Field Fusion (DFF) module and the Progressive Feature Fine (PFF) module, which are applied at each scale.
The DFF is responsible for fusing the predicted multi-scale deformation sub-fields into the current scale via dense propagation. 
The PFF integrates features progressively from three aspects, (\ie, the warped source-infrared features, fixed pseudo-infrared features, and decoding features of the previous scale), which implicitly models the registered features and the remaining misaligned features.
Both of them collaborate to predict the final deformation fields, thus obtaining the registered infrared image by warping operation.

In addition, to generate fusion images that highlight significant structures and textures in the scene, we develop a Transformer-Conv-based Fusion (TCF) subnetwork to simultaneously capture local features and model long-range feature dependencies. 
The feature extractor of the TCF consists of two separate Transformer-Conv Blocks, engineered to extract informative multi-modality features from the registered infrared and visible images. Internally, each Transformer-Conv module incorporates a Swin-Transformer Block (STB) for modeling global feature representations, complemented by a Residual Convolution Block (RCB) for capturing local features.
To adaptively fuse the salient and meaningful portions within the extracted multi-modality features, we utilize a dual-attention mechanism to enable the selection of relevant features from both the infrared and visible features, considering both the spatial and channel dimensions.
Throughout the training, we jointly optimize the C-MPDR and the TCF to mutually improve their performance each other.
We evaluate the proposed method on synthetic misaligned NIR/IR-VIS datasets and real-world misaligned IR-VIS datasets, and demonstrate its effectiveness and robustness, as shown in Fig.~\ref{fig:first_img}(g)-(i). 
In summary, the main contributions are as follows:
\begin{itemize}
\item We propose a novel multi-modality image fusion framework termed IMF, capable of handling displacements and deformations caused by imaging pipelines, accomplishing the suppression of edge ghosts and the correction of structural distortions in the fused image.
	
\item A multi-scale progressive dense registration scheme is involved. At each scale, a deformation field fusion module (DFF) is responsible for fusing all predicted deformation sub-fields up to the current scale, while a progressive feature fine (PFF) module implicitly models registered and remaining misaligned features. Both complement each other to predict the final deformation fields.
	
\item To ensure high-quality fusion images, a Transformer-Conv-based fusion subnetwork is developed to capture local and long-range feature dependencies, emphasizing informative and meaningful information in fused images.
	
\item A data synthesis method that employs hybrid deformations to simulate real-world misalignments between multi-modality images is introduced, allowing us to scale up sufficient training data.
\end{itemize}

The rest of this paper is organized as follows. Sec.~\ref{related_work} reviews
related works. Sec.~\ref{motivation} outlines our motivation. Sec.~\ref{method} details the proposed method. Sec.~\ref{experiment} provides experimental results and effectiveness analysis. Finally, Sec.~\ref{conclusion} concludes this work.

\begin{figure*}[tbh]
    \centering
    \begin{tabular}{c}
	\includegraphics[width=0.95\linewidth]{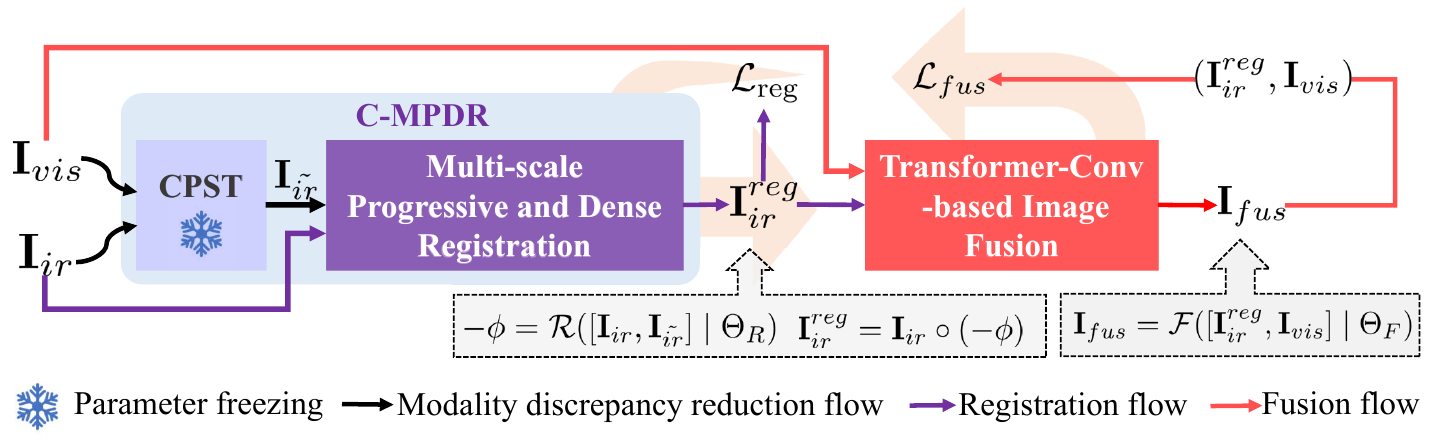} \\
    \end{tabular}
    \vspace{-2mm}
    \caption{Workflow of the proposed IMF for misaligned infrared and visible image fusion, which comprises two main portions. The first portion is a cross-modality multi-scale progressive and dense registration (C-MPDR) network, wherein the frozen CPST in UMF~\cite{UMF} mitigates cross-modality discrepancies and generates a pseudo-infrared image $\mathbf{I}_{\tilde{ir}}$. The MPDR subnetwork registers $\mathbf{I}_{ir}$ with $\mathbf{I}_{\tilde{ir}}$, producing a registered infrared image $\mathbf{I}_{ir}^{reg}$. The second portion is a Transformer-Conv-based fusion (TCF) subnetwork, which fuses $\mathbf{I}_{ir}^{reg}$ and $\mathbf{I}_{vis}$, generating the fused image $\mathbf{I}_{fus}$.
    }%
    \label{fig:overview}
\end{figure*}

\section{Related Work}\label{related_work}
\subsection{Multi-modality Image Fusion}
Early deep learning-based IVIF methods~\cite{DenseFuse, PMGI, FusionDN20, DIDFuse_2020, U2Fusion, MFEIF} adopt a shallow densely connected network~\cite{DenseNet17} to extract multi-modality features since the dense connection can reuse features from each preceding layer to all subsequent layers. 
%
%
%
%
%
%
However, dense networks fall short of retaining intricate high-quality structures and details.
Meanwhile, the employment of generative adversarial networks (GANs) for image fusion has also emerged including FGAN~\cite{FGAN} based on single-discriminator, and DDcGAN~\cite{DDcGAN} and GANMCc~\cite{GANMcC} based on dual-discriminator.
%
%
%
%
With visually pleasing fused images, these hand-crafted designs exhibit structural redundancy as well as a lack of flexibility.
Benefiting from the automatic and differentiable structure discovery capability of the NAS technology, some NAS-based image fusion methods~\cite{SMoA_21, HAFA_21, LSLA_22} have been developed.
%
%
With the Vision Transformers model long-range spatial dependencies, it is natural to apply them to image fusion~\cite{SwinFusion, CGTF, PPTFusion}.
%
In recent two years, several task-oriented image fusion methods~\cite{TarDAL_22, SeAFusion, IRFS_23} have emerged, addressing the issue of one-sided focus on visual quality in image fusion and promoting downstream semantic tasks.

\subsection{Image Registration}
\subsubsection{Mono-modality Image Registration}
Flow-based methods~\cite{flownet, voxelmorph, pwcnet_18, dgcnet_19, glunet_20} have been dominant in estimating dense pixel correspondences between two mono-modality images.
For instance, Dosovitskiy~\etal~\cite{flownet} constructed the first end-to-end network called FlowNet for predicting optical flow and proved the feasibility of estimating flow in a learning manner.
%
%
%
Sun~\etal~\cite{pwcnet_18} proposed learnable feature pyramids instead of traditional image pyramids and designed the warp operation as a layer of the network, as well as utilized a correlation layer to predict the optical flow.
However, these methods perform well for small displacements but may not give good results for strong geometric transformations.
To this end, Melekhov~\etal~\cite{dgcnet_19} built upon the benefits of optical flow estimation and expanded it to large displacements to estimate dense pixel correspondences between two images.
Later, more methods estimating dense flow and correspondences were developed, \ie, GLU-Net~\cite{glunet_20} and GMFlow~\cite{gmflow_22}.
Unfortunately, they are intended to work with mono-modal image pairs and are not suitable for multi-modality data.

\subsubsection{Cross-modality Image Registration}
Research on cross-modality image registration is limited due to the absence of effective similarity measures between different modalities. 
At present, only a handful of registration methods aimed at infrared and visible images have been proposed.
Typically, Arar~\etal~\cite{Nemar_20} developed an unsupervised framework for training multi-modality image translation and spatial registration using two distinct flows. One flow involves registration first, while the other involves translation first. Both flows are implemented simultaneously.
Zhou~\etal~\cite{crossRAFT_22} designed a self-supervised knowledge distillation framework that transfers prior knowledge from mono-modality optical flow models to cross-modality flow estimation.

\subsection{Joint Cross-modality Image Registration and Fusion}
In the past two years, a few fusion methods have emerged for misaligned multi-modality image fusion.
For example, UMF~\cite{UMF} first introduced a cross-modality image registration paradigm that utilizes image-to-image translation to transform the cross-modality image registration problem into a mono-modality problem, thereby rectifying structural distortions and suppressing edge ghosts.
Closest to this work, Xu~\etal~\cite{RFNet_22} proposed a mutually reinforced framework for multi-modality registration and fusion, which divides the registration into a coarse phase and a fine phase, and the fine registration and fusion are optimized jointly. 
%
%
More recently, Tang~\etal~\cite{SuperFusion_22} developed a versatile framework SuperFusion for image registration and fusion, augmented with semantic awareness. It integrates image registration, fusion, and semantic segmentation as a whole. However, this method necessitates ground truth (\ie, accurate deformation fields) for supervision.
%
%

\section{Motivation}\label{motivation}
The variations in imaging pipelines and heat dissipation inside multi-modality sensors result in significant pixel displacement and content deformation between the observed infrared and visible images. This leads to severe structural distortion and edge ghosts in the subsequent multi-modality image fusion.
The mainstream solution is ``registration first, fusion latter". However, existing efforts either rely on separate coarse and fine registrations or directly predict the full-size target deformation fields. The former is not compact and the latter is not accurate enough.
Therefore, we expect to exploit a one-stage coarse-to-fine image registration scheme.
Empirically, within multi-scale features, deep features with low resolution and large receptive fields can characterize global structures, and then coarse registration on deep features is ideal for correcting large displacements.
In contrast, shallow features with high resolution and small receptive fields can characterize local details, and then fine registration on shallow features is better for correcting local deformations.
Motivated by this, we are committed to exploiting a novel one-stage coarse-to-fine image registration model to predict the final deformation fields in a compact and progressively refined manner, thus improving the quality of multi-modality image fusion. 
%


\section{The proposed method}\label{method}

\subsection{Overview}\label{overview}
Fig.~\ref{fig:overview} depicts the overall workflow of our IMF. 
Given a pair of misaligned infrared and visible images, its goal is to register the infrared image $\mathbf{I}_{ir}$ onto the visible image $\mathbf{I}_{vis}$ thus output a fusion image $\mathbf{I}_{fus}$ without structural distortions and edge ghosts.
To accomplish this goal, two important procedures are involved:
1) cross-modality image registration by our multi-scale progressive and dense registration subnetwork equipped with the pioneer CPST proposed by~\cite{UMF}, called C-MPDR;
2) cross-modality image fusion by the transformer-conv-based fusion subnetwork, called TCF.
In Fig.~\ref{fig:overview}, the black, purple, and red arrows represent the flows of the modality discrepancy reduction flow, near mono-modality registration flow, and fusion flow, respectively.
Specifically, in the C-MPDR, the CPST first translates a visible image into a pseudo-infrared image $\mathbf{I}_{\tilde{ir}}$.
Next, $\mathbf{I}_{\tilde{ir}}$ is treated as the fixed image, and then source distorted infrared image $\mathbf{I}_{ir}$ is warped onto it through the MPDR in a near mono-modality registration pattern, resulting in registered infrared image $\mathbf{I}_{ir}^{reg}$.
Finally, TCF extracts local and global multi-modality features from the $\mathbf{I}_{ir}^{reg}$ and $\mathbf{I}_{vis}$ via the Transformer-Conv~\cite{scunet_22} blocks and then fuses them into an image with salient and meaningful information.
During the training, the C-MPDR and TCF subnetworks are trained jointly, while the CPST remains frozen. The circular arrows between C-MPDR and TCF signify the bidirectional transmission of image content and gradients.

\begin{figure*}[t]
	\centering
	\begin{tabular}{c}
		\includegraphics[width=0.95\linewidth]{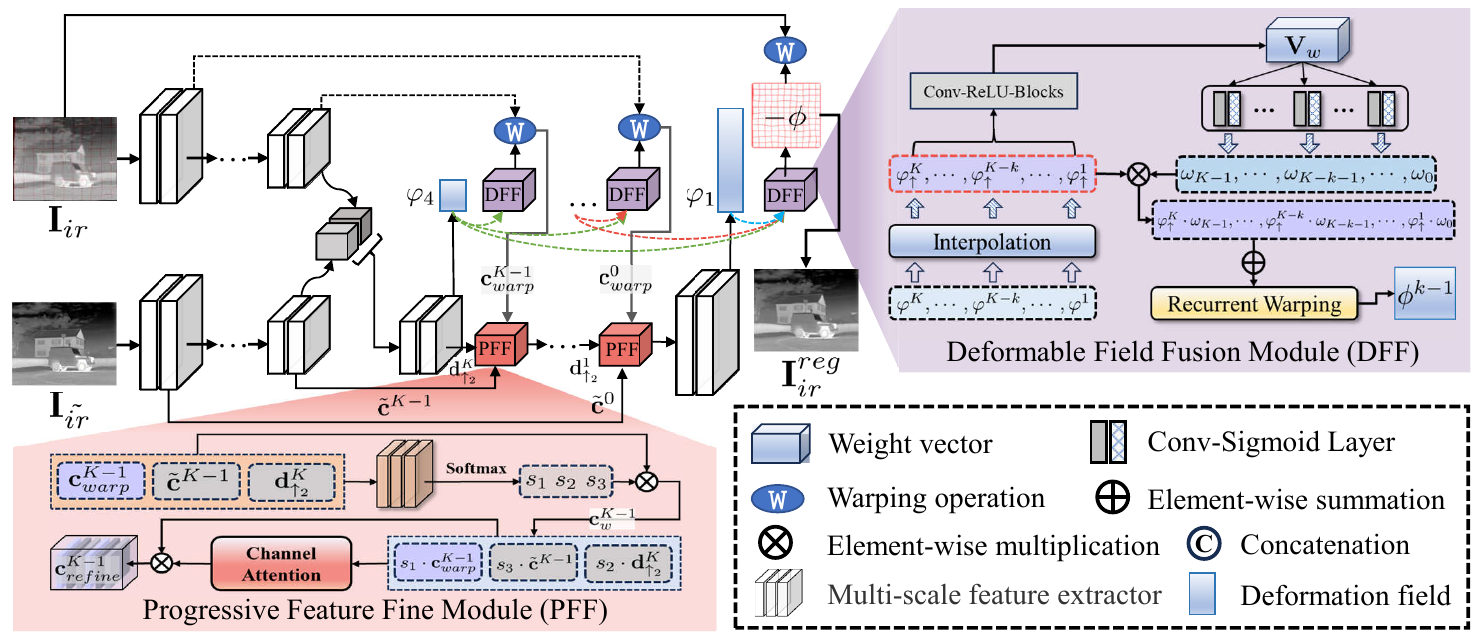} \\
	\end{tabular}
	\caption{Schematic of the proposed multi-scale progressive and dense registration (MPDR) subnetwork. Here, two independent feature extractors are utilized to extract hierarchical features from distorted-infrared and pseudo-infrared images, respectively. Subsequently, two core components, namely the deformable field fusion module (DFF) and the progressive feature fine module (PFF), collaborate to predict multi-scale deformation subfields in a dense propagation manner.}  %
	\label{fig:MPDR_Detail}
\end{figure*}

\subsection{Multi-scale Progressive and Dense Registration}\label{mpdr}
Thanks to the reduction of cross-modality discrepancy by CPSTN, the problem of multi-modality image registration is simplified to near mono-modality registration.
As shown in Fig.~\ref{fig:MPDR_Detail}, we exploit a Multi-scale Progressive and Dense Registration subnetwork (MPDRN) to estimate the deformation fields $-\phi$ between source distorted infrared image $\mathbf{I}_{ir}$ and the pseudo-infrared image $\mathbf{I}_{\tilde{ir}}$, and then warp $\mathbf{I}_{ir}$ onto $\mathbf{I}_{\tilde{ir}}$ with $-\phi$, thus obtaining the registered infrared image $\mathbf{I}_{ir}^{reg}$. Overall, the registration process can be roughly represented as
\begin{equation}
	\begin{split}
		\label{eq:register}
		-\phi = \mathcal{R}\left(\mathbf{I}_{\tilde{ir}},\mathbf{I}_{ir}\mid\Theta_{R}\right),
		\mathbf{I}_{ir}^{reg} = \mathbf{I}_{ir} \circ (-\phi),
	\end{split}
\end{equation}		 
where $\mathcal{R\left(\cdot\right)}$ refers to the MPDRN with parameter $\Theta_{R}$ and the symbol $\circ$ denotes the warping operation.

Specifically, the MPDRN consists of four key components, \ie, a learnable two-stream pyramid feature extractor, a multi-scale decoder, and a series of Deformable Field Fusion (DFF) and Progressive Feature Fine (PFF) modules, as shown in Fig.~\ref{fig:MPDR_Detail}.
Firstly, to capture individual abundant feature representations from $\mathbf{I}_{\tilde{ir}}$ and $\mathbf{I}_{ir}$, we abandon the shared two-scale feature extractor in~\cite{UMF} and replace it with a learnable two-stream feature pyramid.
Through it, two $K-$scale feature pyramids are extracted from $\mathbf{I}_{\tilde{ir}}$ and $\mathbf{I}_{ir}$, denoted as $\left[\tilde{\mathbf{c}}^{0}, \cdots, \tilde{\mathbf{c}}^{k}, \cdots, \tilde{\mathbf{c}}^{K}\right]$ and $\left[\mathbf{c}^{0}, \cdots, \mathbf{c}^{k}, \cdots, \mathbf{c}^{K}\right]$, respectively. In this work, $K$ is set to $4$.
At the top scale, $\tilde{\mathbf{c}}^{K}$ and $\mathbf{c}^{K}$ are first integrated by concatenation, and then fed into the top scale of the multi-scale decoder to obtain decoded features $\mathbf{d}^{K}$ and estimate the deformation fields $\varphi^{K}$. 
Then, the $\varphi^{K}$ is appended to the DFF module. Here, since $\varphi^{K}$ is the only one, it undergoes a self-integration and becomes $\phi^{K-1}$ in the DFF module. we directly warp the $(K-1)$-\textit{th} scale features from the source distorted infrared image toward the pseudo-infrared image using $\phi^{K-1}$:
\begin{equation}
	\label{eq:warp}
	\mathbf{c}_{warp}^{K-1}(\mathbf{p})=\mathbf{c}^{K-1}\left(\mathbf{p}+\phi^{K-1}(\mathbf{p})\right),
\end{equation}
where $\mathbf{p}$ refers to the position index.
The warped features $\mathbf{c}_{warp}^{K-1}$, the upsampled decoded features $\mathbf{d}_{\uparrow_2}^{K}$, and the $(K-1)$-\textit{th} scale features $\tilde{\mathbf{c}}^{K-1}$ from the pseudo-infrared image are integrated and refined through the PFF module:
\begin{equation}
	\label{eq:refine}
	\mathbf{c}_{refine}^{K-1}=PFF\left([\mathbf{c}_{warp}^{K-1},\mathbf{d}_{\uparrow_2}^{K},\tilde{\mathbf{c}}^{K-1}]\right),
\end{equation}
where $[\cdot,\cdot]$ refers to the concatenation, the symbol $\uparrow_{2}$ denotes the $2\times$ upsampling operation.
Following Eq.~(\ref{eq:warp}) and Eq.~(\ref{eq:refine}), we sequentially obtain multi-scale fused deformation fields$\left[\phi^{K}, \cdots, \phi^{k}, \cdots, \phi^{0}\right]$ by the DFF modules, warped feature representations $\left[\mathbf{c}_{warp}^{K}, \cdots, \mathbf{c}_{warp}^{k}, \cdots, \mathbf{c}_{warp}^{0}\right]$ of the source distorted infrared image, and progressively refined features $\left[\mathbf{c}_{refine}^{K}, \cdots, \mathbf{c}_{refine}^{k}, \cdots, \mathbf{c}_{refine}^{0}\right]$ by the PFF modules.
Next, we will explain the main ideas for two core components of the MPDRN in detail~\ie, DFF, and PFF.

\subsubsection{\bf{Deformable Field Fusion (DFF) Module}}
Due to the practical challenges in directly predicting the targeted full-size deformation fields between the pseudo-IR image $\mathbf{I}_{\tilde{ir}}$ and the source distorted-IR image $\mathbf{I}_{ir}$, we intend to treat the prediction of the target deformation fields as a coarse-to-fine process.
As shown in Fig.~\ref{fig:MPDR_Detail}, we continuously reuse the coarse deformation fields predicted at each previous scale in a dense propagation manner until the final deformation fields $-\phi$ are estimated. Here, the proposed DFF module is responsible for the reuse of multi-scale deformation fields.
As illustrated in the enlarged purple window in Fig.~\ref{fig:MPDR_Detail}, distinct from addition or concatenation, the DFF utilizes reweighting to fuse the coarse deformation fields with different scales.
Suppose that the coarse deformation fields $\varphi^{k}$ of the $k$-th scale has been estimated, and previous $\varphi^{K}, \cdots, \varphi^{k-1}$ already existed. Then, we first interpolate them to a size as large as $\varphi^{k}$ by
\begin{equation}
	\label{eq:interpolation}
	\varphi_{\uparrow}^{i}=\mathcal{P}_{2}^{i-k}\left(\varphi^{i}\right),
\end{equation}
where $i \in [K, K-1, \cdots, k]$ and $\mathcal{P}_{2}^{i-k}\left(\cdot\right)$ denotes the bilinear interpolation function with scale factor $2^{i-k}$.
Next, $\varphi_{\uparrow}^{K}, \varphi_{\uparrow}^{K-1}, \cdots, \varphi_{\uparrow}^{k}$ are reweighted. Specifically, a total weighting vector for those interpolated deformation fields is calculated by
\begin{equation}
	\label{eq:total_weights}
	\mathbf{V}_{w}=\mathcal{M}_{\texttt{CRB}}\left(\left[\varphi_{\uparrow}^{K}, \varphi_{\uparrow}^{K-1}, \cdots, \varphi_{\uparrow}^{k}\right]\right),
\end{equation} 
where $\mathcal{M}_{\texttt{CRB}}\left(\cdot\right)$ is a mapping from the merged deformation fields by $\left[\cdot, \cdot\right]$ into the weight vector $\mathbf{V}_{w}$. \texttt{CRB} refers to Conv-ReLU Blocks.
The DFF module then assigns weights to multi-scale coarse deformation fields that were estimated earlier by
\begin{equation}
	\label{eq:each_weight}
	\mathbf{\omega}_{t}=\mathcal{M}_{\texttt{CSB}}\left[t\right]\left(\mathbf{V}_{w}\right),
\end{equation} 
where $\mathcal{M}_{\texttt{CSB}}\left[t\right]\left(\cdot\right)$ denotes a series of mapping functions from the $\mathbf{V}_{w}$ into each deformation fields $\varphi_{\uparrow}^{i}$, and they consist of a single Conv-Sigmoid block. $t$ denotes the index, and $t \in \left[K-k, \cdots, 1, 0\right]$. 
Immediately, we reweight these deformation fields after interpolation using a simple weighted summation. and then follow a vector integration~\cite{voxelmorph} calculation to implement recurrent warping, thus obtaining the refined deformation fields as:
\begin{equation}
	\label{eq:refined}
	\phi^{k-1}=\mathcal{V}\left(\sum_{i=k}^{K}\sum_{t=0}^{K-k}\mathbf{\omega}_{t} \otimes \varphi_{\uparrow}^{i}\right),
\end{equation} 
where $\mathcal{V}\left(\cdot\right)$ denotes vector integration and $\otimes$ denotes the element-wise multiplication. When $k=1$, it means that the requested refined deformation fields between the pseudo-IR and source distorted-IR images are estimated,~\ie, $-\phi=\phi^{0}$.
Returning $-\phi$ to Eq.~(\ref{eq:register}), a registered-IR image is generated.

\subsubsection{\bf Progressive Feature Fine (PFF) Module}
The DFF only reuses the previous multi-scale deformation fields, but the accurate estimation of deformation fields is inseparable from the required feature representations.
To this end, we design another module termed PFF, which progressively aggregates features that are beneficial for estimating the more refined deformation fields at the current scale.
The so-called "progressively" means that the inherent offset between the pseudo-IR image and the source distorted-IR image is greatly corrected at the earlier scales, while the current PFF block focuses on estimating more refined deformation fields for any remaining offset.
As illustrated in the enlarged red window in Fig.~\ref{fig:MPDR_Detail}, taking the $(K-1)$-\textit{th} scale as an example, the first step involves merging features from three parts,~\ie, $\mathbf{c}_{warp}^{K-1}$, $\mathbf{d}_{\uparrow_2}^{K}$, and $\tilde{\mathbf{c}}^{K-1}$ as expressed in Eq.~(\ref{eq:refine}) through concatenation and Conv-ReLU blocks. We mark the merged feature as $\mathbf{c}_{mix}^{K-1}$.
During the following stage of feature refinement, we first calculate the weights for the three input parts by
\begin{equation}
	\label{eq:weight}
	\begin{split}
		s_{1},s_{2},s_{3}=\texttt{Softmax}\left(\mathcal{M}_{\texttt{CRB}}(\mathbf{c}_{mix}^{K-1})\right).
	\end{split}
\end{equation} 
As a result, we obtain the weighted merging features through $\mathbf{c}_{w}^{K-1}=\left[s_{1}\cdot\mathbf{c}_{warp}^{K-1},s_{2}\cdot\mathbf{d}_{\uparrow_2}^{K}, s_{3}\cdot\tilde{\mathbf{c}}^{K-1}\right]$.
Next, we refine the weighted feature leveraging a channel attention~\cite{CA_18} branch. This process can be expressed as
\begin{equation}
	\label{eq:pff}
	\begin{split}
		\mathbf{c}_{refine}^{K-1}=\mathbf{c}_{w}^{K-1}\otimes\left(\mathcal{A}_{\texttt{channel}}\left(\mathbf{c}_{w}^{K-1}\right)\right),
	\end{split}
\end{equation} 
where $\mathcal{A}_{\texttt{channel}}\left(\cdot\right)$ refers to the channel attention module. The $\mathbf{c}_{refine}^{K-1}$ is the refined features at the $(K-1)$-\textit{th} scale.

\subsubsection{\bf Loss functions for Registration}\label{subsec:reg loss}
To constrain the registration of distorted IR and pseudo-IR images at feature level, we introduce a bidirectional similarity loss that is defined as follows
\begin{equation}
	\begin{split}
		\mathcal{L}_{sim}^{bi}&=\|\psi_{j}(\mathbf{I}_{ir}^{reg})-\psi_{j}(\mathbf{I}_{\tilde{ir}}) \|_{1} \\ &+ \lambda_{rev} \|\psi_{j}(\phi \circ \mathbf{I}_{\tilde{ir}})-\psi_{j}(\mathbf{I}_{ir})\|_{1},
	\end{split}
\end{equation}
where the first term is forward, while the second term is backward with weight $\lambda_{rev}=0.2$. In the second term, we use the reverse deformation fields $\phi$ to distort the pseudo-IR image $I_{\tilde{ir}}$ and bring it closer to the source distorted-IR input $I_{ir}$.
To ensure smooth deformation fields, we introduce a smooth loss, defined as follow
\begin{equation}
	\mathcal{L}_{\text {smooth }}=\|\nabla \phi\|_{1}.
\end{equation}
Finally, the overall loss for registration is defined as
\begin{equation}
	\label{eq:total_reg_loss}
	\mathcal{L}_{\mathrm{reg}}=\mathcal{L}_{sim}+\lambda_{sm} \mathcal{L}_{smooth},
\end{equation}
where the trade-off parameter $\lambda_{sm}$ is set to $10$ in this work.

\begin{figure}[t]
	\centering
	\begin{tabular}{c}
		\includegraphics[width=0.95\linewidth]{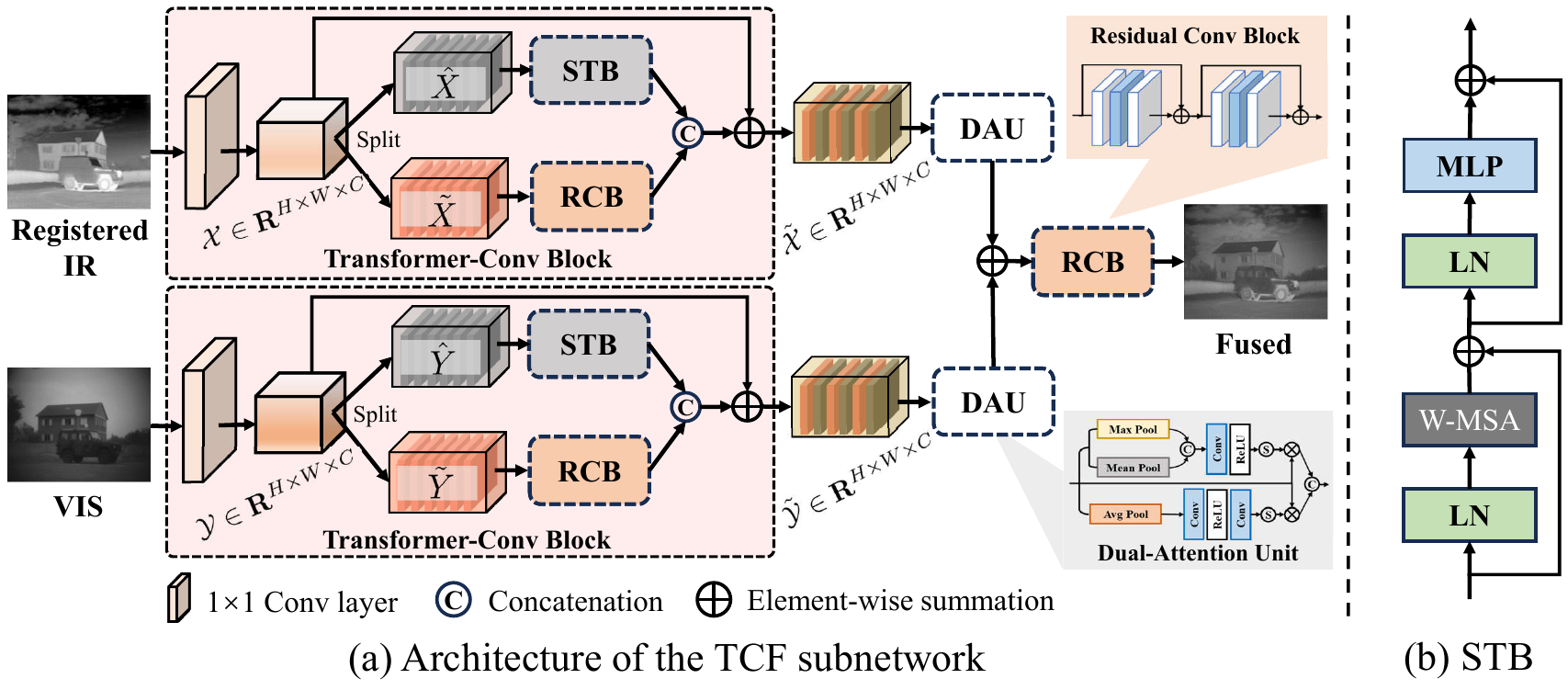} \\
	\end{tabular}
	\caption{(a) The architecture of the Transformer-Conv-based image fusion subnetwork (TCFN). (b) Swin Transformer block. Noted that W-MSA refers to the multi-head self-attention mechanism with shifted windowing configuration.}  %
	\label{fig:TCF}
\end{figure}

\subsection{Transformer-Conv-based Image Fusion}\label{tcf}
To fuse registered-IR image and visible image, we develop a Transformer-Conv-based image fusion subnetwork (TCFN).
Different from pioneer IVIF methods~\cite{DenseFuse,DDcGAN,PMGI} that commonly use DenseBlock~\cite{DenseNet17} as feature extractors, our TCFN introduces two swinTransformer-Conv Blocks (TCB) inspired by~\cite{scunet_22} to model local and non-local features of infrared and visible images.
The architecture of the TCFN is shown in Fig.~\ref{fig:TCF}(a). It involves two separate streams that rely on TCBs for extracting multi-modality features, a fusion strategy that incorporates Dual-Attention Units (DAU) and element-wise summation, and two residual convolutional blocks.
Given two inputs $\mathbf{I}_{ir}^{reg}$ and $\mathbf{I}_{vis}$, their overall fusion process is formulated as
\begin{equation}
	\mathbf{I}_{fus} = \mathcal{F}\left(\mathbf{I}_{ir}^{reg},\mathbf{I}_{vis} \mid \Theta_{F} \right),
\end{equation}
where $\mathcal{F}\left(\cdot\right)$ is the TCFN, $\Theta_{F}$ denotes its parameters. 
We then detail two key components of the TCDN~\ie, TCB, and DAU.

\subsubsection{\bf swinTransformer-Conv Blocks (TCB)}
Fig.~\ref{fig:TCF}(a) illustrates that the TCB is also a residual structure that comprises a $1\times1$ convolutional layer, a SwinTransformer-Block (STB)~\cite{swinIR_21}, and a Residual-Conv-Block (RCB). 
Specifically, the $\mathbf{I}_{ir}^{reg}$ and $\mathbf{I}_{vis}$ first passe through the $1\times1$ convolutional layer thus generate features $\mathcal{X} \in \mathbb{R}^{H \times W \times C}$ and $\mathcal{Y} \in \mathbb{R}^{H \times W \times C}$, respectively.
Subsequently, $\mathcal{X}$ and $\mathcal{Y}$ are split into two groups of feature maps \ie, $[\hat{X}, \tilde{X}]$ and $[\hat{Y}, \tilde{Y}]$. This process is expressed as
\begin{equation}
	\begin{split}
		\hat{\mathbf{X}}, \tilde{\mathbf{X}} = \texttt{Split}\left(\texttt{Conv}_{1\times1}\left(\mathcal{X}\right)\right),\\
		\hat{\mathbf{Y}}, \tilde{\mathbf{Y}} = \texttt{Split}\left(\texttt{Conv}_{1\times1}\left(\mathcal{Y}\right)\right).
	\end{split}
\end{equation}
For the infrared branch, $\hat{X}$ is then input into the STB to capture long-range global feature dependencies, while $\tilde{X}$ is processed by the RCB to learn local feature dependencies. After that, concatenation and skip connection are applied to generate the global-local feature mixture $\tilde{\mathcal{X}}$. The visible branch generates $\tilde{\mathcal{Y}}$ in a similar approach. As such, the outputs of the TCBs are formulated as
\begin{equation}
	\begin{split}
		\tilde{\mathcal{X}} &=\texttt{Conv}_{1\times1}([\texttt{SwinT}(\hat{\mathbf{X}}),\texttt{RConv}(\tilde{\mathbf{X}})])+\mathcal{X},\\
		\tilde{\mathcal{Y}} &=\texttt{Conv}_{1\times1}([\texttt{SwinT}(\hat{\mathbf{Y}}),\texttt{RConv}(\tilde{\mathbf{Y}})])+\mathcal{Y},
	\end{split}
\end{equation}
where $\texttt{RConv}\left(\cdot\right)$ denotes the RCB and $\texttt{SwinT}\left(\cdot\right)$ denotes the TCB. 
As shown in Fig.~\ref{fig:TCF}(b), the TCB mainly contains two phases, \ie, multi-head self-attention (MSA) and multi-layer perceptron (MLP).  Here, two LayerNorm (LN) layers are respectively added before both MSA and MLP, and two skip connections are employed for both phases. Let $\mathbf{Z} \in \mathbb{R}^{H \times W \times C}$ denote the input to the TCB, this process is formulated as
\begin{equation}
	\left\{\begin{array}{l}
		\mathbf{Z} = MSA\left(LN\left(\mathbf{Z}\right)\right)+\mathbf{Z},\\
		\mathbf{Z} = MLP\left(LN\left(\mathbf{Z}\right)\right)+\mathbf{Z}.
	\end{array}\right.
\end{equation}
\subsubsection{\bf Dual-Attention Units (DAU)}
To aggregate content-rich features from $\tilde{\mathcal{X}}$ and $\tilde{\mathcal{Y}}$ in both channel and spatial dimensions, we employ a dual-attention-based fusion strategy to first enhance the multi-modality feature representations before merging them.
Fig.~\ref{fig:TCF}(a) shows the DAU in an enlarged gray window, which embraces two parallel branches, one for spatial attention~\cite{SA_18} and the other for channel attention~\cite{CA_18} of multi-modality $\tilde{\mathcal{X}}$ and $\tilde{\mathcal{Y}}$.
As such, the enhanced infrared-modality features $\tilde{\mathcal{X}}_{\texttt{EN}}$ are obtained by the following process
\begin{equation}
	\left\{\begin{array}{l}
		\tilde{\mathcal{X}}_{\texttt{SA}} =  \texttt{Sigmoid}(\texttt{CR}([\texttt{MaxP}(\tilde{\mathcal{X}}),\texttt{MeanP}(\tilde{\mathcal{X}})]))\otimes \tilde{\mathcal{X}},\\
		\tilde{\mathcal{X}}_{\texttt{CA}} =  \texttt{Sigmoid}(\texttt{CRC}(\texttt{AvgP}(\tilde{\mathcal{X}})))\otimes \tilde{\mathcal{X}},\\
		\tilde{\mathcal{X}}_{\texttt{EN}} = \texttt{CR}([\tilde{\mathcal{X}}_{\texttt{SA}},\tilde{\mathcal{X}}_{\texttt{CA}}]),
	\end{array}\right.
\end{equation}
where $\texttt{CR}\left(\cdot\right)$ denotes the Conv-ReLU layers, $\texttt{CRC}\left(\cdot\right)$ denotes the Conv-ReLU-Conv layers.
The $\texttt{MaxP}\left(\cdot\right)$, $\texttt{MeanP}\left(\cdot\right)$, and $\texttt{AvgP}\left(\cdot\right)$ represent the max pooling layer, mean pooling layer, and global average pooling layer, respectively.
A similar process is applied to enhance the visible-modality features $\tilde{\mathcal{Y}}$, resulting in $\tilde{\mathcal{Y}}_{\texttt{EN}}$.
Finally, we merge these enhanced multi-modality features and generate the fused image by
\begin{equation}
	\mathbf{I}_{fus} = \texttt{Conv}_{3\times 3}(\texttt{RConv}(\tilde{\mathcal{X}}_{\texttt{EN}} + \tilde{\mathcal{Y}}_{\texttt{EN}})).
\end{equation}

\subsubsection{\bf Loss functions for Fusion}\label{subsec:fusion loss}
We utilize MS-SSIM loss function $\mathcal{L}_{ssim}^{ms}$ to maintain sharper intensity distribution of the fused image, which is defined as
\begin{equation}
	\mathcal{L}_{ssim}^{ms}=\left(1-SSIM(\mathbf{I}_{fus}, \mathbf{I}_{ir}^{reg})\right)+\left(1-SSIM(\mathbf{I}_{fus}, \mathbf{I}_{vis})\right).
\end{equation}
\hspace{-1mm}To encourage the restoration of texture details, we model the gradient distribution and develop a joint gradient loss as
\begin{equation}
	\mathcal{L}_{JGrad}=\|max\left(\nabla \mathbf{I}_{ir}^{reg},\nabla \mathbf{I}_{vis}\right), \nabla \mathbf{I}_{fus} \|_{1}.
\end{equation}
For clearer textures, the gradient of the fused image is forced to approach the maximum value between the infrared and visible image gradients. 
In addition, to retain the saliency targets from the two images, we leverage self-visual saliency maps inspired by~\cite{SvSW,SMoA_21,TarDAL_22} to build a new loss as 
\begin{equation}
	\begin{split}
		\boldsymbol{\omega}_{ir}&=\boldsymbol{S}_{f_{i r}} /\left(\boldsymbol{S}_{f_{i r}}-\boldsymbol{S}_{f_{v i s}}\right),
		\boldsymbol{\omega}_{vis}=1-\boldsymbol{\omega}_{ir}, \\
		\mathcal{L}_{svs}&=\|\left(\boldsymbol{\omega}_{ir} \otimes \mathbf{I}_{ir}^{reg}+\boldsymbol{\omega}_{vis} \otimes \mathbf{I}_{vis}\right), I_{fus} \|_{1},
	\end{split}
\end{equation}
where $\boldsymbol{S}$ denotes the saliency matrix, $\boldsymbol{\omega}_{vis}$ and $\boldsymbol{\omega}_{ir}$ denote the weight maps for infrared and visible images, respectively.
The overall fusion loss is computed by
\begin{equation}
	\label{eq:total_fusion_loss}
	\mathcal{L}_{fus}=\lambda_{ssim}\mathcal{L}_{ssim}^{ms} + \lambda_{JG}\mathcal{L}_{JGrad} + \lambda_{svs}\mathcal{L}_{svs},
\end{equation}
where $\lambda_{ssim}$, $\lambda_{JG}$, and $\lambda_{svs}$ are set to $1.0$, $20.0$, and $5.0$.

\section{Experiments}\label{experiment}

\subsection{Datasets}
We train and validate the proposed IMF on misaligned datasets including:
\subsubsection{\bf{Synthetic misaligned IR/NIR-VIS images}}
These images are generated by simulating the visual distortions caused by camera position deviation, high temperature of CMOS in the camera, and thermal diffusion of objects in the hot environment, on TNO\footnote{http://figshare.com/articles/TNO\_Image\_Fusion\_Dataset/1008029.}, RoadScene\footnote{https://github.com/hanna-xu/RoadScene.}, and M$^3$FD~\cite{TarDAL_22}, three infrared and visible (IR-VIS) datasets, and a near-infrared and visible (NIR-VIS) dataset MSIFT~\cite{MSIFT}, which are originally well-aligned. For the formation of misaligned image pairs during training, we first generate random affine transformations by adjusting $7$ transformation parameters to simulate geometric deformation. Given one rotation $\theta$, two translations $\left(\text{t}_{x}, \text{t}_{y}\right)$, two scalings $\left(\text{c}_{x}, \text{c}_{y}\right)$, and two shears $\left(\text{s}_{x}, \text{s}_{y}\right)$, we obtain the affine transformation matrix $\textbf{M}$ according to
\begin{equation}
	\begin{small}
		\left[\begin{array}{ccc}
			\text{c}_{x}\cos(\theta)+\text{s}_{x} \text{c}_{x}\sin(\theta) & -\text{c}_{x}\sin(\theta)+\text{s}_{x} \text{c}_{x}\cos(\theta) & \text{t}_{x} \\
			\text{s}_{y} \text{c}_{y}\cos(\theta)+\text{c}_{y} \sin(\theta) & -\text{s}_{y} \text{c}_{y} \sin(\theta)+\text{c}_{y} \cos(\theta) & \text{t}_{y} \\
			0 & 0 & 1
		\end{array}\right].
	\end{small}	
\end{equation}
After that, we warp the original infrared image $I_{ir}$ by the $\textbf{M}$ to generate the affine-distorted image as follows
\begin{equation}
	\begin{small}
		I_{\texttt{affine}} = \mathcal{W}_{\texttt{affine}}(I_{ir},\textbf{M}),
	\end{small}	
\end{equation}
where $\mathcal{W}(\cdot)$ denotes the warping operation. 
%
%
And then, we proceeded to apply elastic deformations to $I_{\texttt{affine}}$ to simulate deformable distortion. Following~\cite{elastic}, we create random displacement fields, that is $\Delta \textbf{x}(x,y)=\operatorname{rand}(-1,+1)$ and $\Delta \textbf{y}(x, y)=\operatorname{rand}(-1,+1)$, where $\operatorname{rand}(-1,+1)$ follows a uniform distribution and $(x,y)$ indexes each pixel of the image. Next, $\Delta \textbf{x}$ and $\Delta \textbf{y}$ are convolved with a Gaussian filtering $\mathcal{G}$ of standard deviation $\sigma$ to generate elastic deformation fields as follows
\begin{equation}
	\begin{small}
		\widetilde{\Delta \textbf{x}}, \widetilde{\Delta \textbf{y}} = \mathcal{G}([\Delta \textbf{x}, \Delta \textbf{y}], \sigma),
	\end{small}	
\end{equation}
where $\sigma$ is regarded as the elasticity coefficient. To control different intensities of deformation at different local regions of the image, the deformation fields $\widetilde{\Delta \textbf{x}}$ and $\widetilde{\Delta \textbf{y}}$ are then multiplied with a scaling factor $\alpha$ which is a positive constant. Thus, the deformable displacement fields are denoted by $\alpha \widetilde{\Delta \textbf{x}}$ and $\alpha \widetilde{\Delta \textbf{y}}$. Finally, we warp the affine-distorted image $I_{\texttt{affine}}$ by $\alpha \widetilde{\Delta \textbf{x}}$ and $\alpha \widetilde{\Delta \textbf{y}}$ to generate the desired misaligned image with a hybrid deformation as follows
\begin{equation}
	\begin{small}
		I_{\texttt{hybrid}} = \mathcal{ W}_{\texttt{elastic}}(I_{\texttt{affine}},(\alpha  \widetilde{\Delta \textbf{x}}, \alpha \widetilde{\Delta \textbf{y}})).
	\end{small}	
\end{equation}
%
%

Regarding the parameter settings, we intend to simulate three different degrees of deformation for infrared images,~\ie, slight, moderate, and severe deformations. A detailed description of parameter settings is provided in Table~\ref{tab:deformable_parameters}.
Noted that all experiments conducted in this work are based on misaligned data with slight deformations, whereas misaligned data with moderate and severe deformations are employed to evaluate the robustness of the proposed method.

\begin{table}[t]
	\caption{Detailed descriptions of parameters of different types of deformations. }
	\label{tab:deformable_parameters}
	\vspace{-6mm}
	\begin{center}
		\renewcommand\arraystretch{1.4} 
		\setlength{\tabcolsep}{6.0pt}{ 
			\resizebox{1.0\linewidth}{!}{
				\begin{tabular}{lc}
					\toprule[1.5pt]
					\textbf{Type}&\textbf{Description of Parameters}\\
					\midrule
					\specialrule{0em}{1pt}{1pt}
					\multirow{2}{*}{\textit{Slight (L)}} & \makecell{\textbf{\textit{Affine}}$-$ rotation: None; translation: $\text{t}_{x},\text{t}_{y} \in\left[-0.01, 0.01\right]$; ~~~~~~~~~~~~ \\scaling: None; shearing: None;}\\
					& \makecell{\textbf{\textit{Deformable}}$-$ displacement: $\Delta \textbf{x}, \Delta \textbf{y}=\operatorname{rand}(-1,+1)$; standard~~~~~~~~\\ deviation: $\sigma \in \left[24,32\right]$; scale factor: $\alpha=1.0$;}\\
					\midrule
					
					\multirow{2}{*}{\textit{Moderate (M)}} & \makecell{\textbf{\textit{Affine}}$-$ rotation: $\theta \in \left[-5^{\circ}, 5^{\circ}\right]$; translation: $\text{t}_{x},\text{t}_{y} \in\left[-0.02, 0.02\right]$;~~~\\ scaling: None; shearing: None;}\\
					& \makecell{\textbf{\textit{Deformable}}$-$ displacement: $\Delta \textbf{x}, \Delta \textbf{y}=\operatorname{rand}(-1,+1)$; standard~~~~~~~~~~\\ deviation: $\sigma \in \left[16,32\right]$; scale factor: $\alpha \in \left[1.0,1.2\right]$;}\\
					\midrule
					
					\multirow{2}{*}{\textit{Severe (S)}} & \makecell{\textbf{\textit{Affine}}$-$ rotation: $\theta \in \left[-10^{\circ}, 10^{\circ}\right]$; translation: $\text{t}_{x},\text{t}_{y} \in\left[-0.02, 0.05\right]$;\\ scaling: None; shearing: None;}\\
					& \makecell{\textbf{\textit{Deformable}}$-$ displacement: $\Delta \textbf{x}, \Delta \textbf{y}=\operatorname{rand}(-1,+1)$; standard~~~~~~~~~~~\\ deviation: $\sigma \in \left[16,24\right]$; scale factor: $\alpha \in \left[1.0,1.2\right]$;}\\	
					\bottomrule[1.5pt]
		\end{tabular}}}
	\end{center}
	\vspace{-4mm}
\end{table}

\subsubsection{\bf{Real-world misaligned IR-VIS images}}
They are from the large-scale FLIR dataset\footnote{https://www.flir.com/oem/adas/adas-dataset-form/.}, which was captured using a vehicle-mounted camera that integrates both RGB and thermal sensors. One subset contains $10,228$ infrared-visible image pairs sampled from short videos. Due to external factors such as road bumps and internal factors such as heat inside sensors, this subset contains quite a few misaligned visible and infrared image pairs. We selected $477$ image pairs with obvious misalignment from it to validate our IMF.

\subsection{Evaluation Metrics} 
\subsubsection{\bf{Registration Metrics}}
we evaluate the registered results using three metrics, \ie, MSE, Mutual Information (MI)~\cite{MI_metric}, and Normalized Cross Correlation (NCC)~\cite{NCC_metric}.
MSE presents the mean square error of two images. MI measures the independence of random variables in the information theory and is also a widely-used similarity metric in image registration and fusion. NCC measures the similarity between matching windows in two images.
A well-aligned pair of images is indicated by a lower MSE, higher MI, and higher NCC.

\subsubsection{\bf{Fusion Metrics}}
we evaluate the fused images using Cross Correlation (CC), Mutual Information (MI)~\cite{MI_metric}, Visual Information Fidelity (VIF)~\cite{vif}, Structural Similarity Index (SSIM)~\cite{ssim}, and \text{Q}$_{abf}$.
MI measures the total amount of information propagated from source images to the fused image. 
VIF measures the information fidelity of the fused images. 
\text{Q}$_{abf}$ measures the total edge information transferred from the source images to the fused image.

\subsection{Implementation Details}
The proposed framework is implemented in PyTorch.
During the training phase, we use all images of RoadScene, M$^3$FD, and MSIFT datasets and randomly select $8$ image patches of size $256\times256$ to form a batch. 
During the testing phase, we select $121$ images from the RoadScene dataset, $24$ images from the TNO dataset, and all images from M$^3$FD and MSIFT datasets as testing data, respectively.
We use a single NVIDIA 1080Ti GPU for training and testing. The Adam optimizer ($\beta _{1}=0.9$, and $\beta _{2}=0.999$) is used to optimize our model.
The initial learning rate is set to $0.001$ and remains unchanged throughout the training phase that goes through $300$ epochs.
The registration loss in Eq.~(\ref{eq:total_reg_loss}) and fusion loss in Eq.~(\ref{eq:total_fusion_loss}) exert equal constraints on the framework.
%
%
%

\begin{table*}[t]
	\begin{center}
		\caption{Quantitative evaluation of cross-modality image registration on the TNO, Roadscene, M$^3$FD~\cite{TarDAL_22}, and MSIFT~\cite{MSIFT} datasets. The top three results are highlighted using \textcolor{myred}{\textbf{red}}, \textcolor{myblue}{\textbf{blue}} and \textcolor{mygreen}{\textbf{green}} in the order. "$\dagger$" denotes that the \textit{\bf CPST}~\cite{UMF} is treated as a plugin to several state-of-the-art mono-modality optical flow models to implement cross-modality flow estimation and image registration.}
		\label{tab:align-resluts}
		\vspace{-4mm}
		\renewcommand
		\arraystretch{1.3} 
		\setlength{\tabcolsep}{6pt}{ 
			\resizebox{1.0\linewidth}{!}{
				\begin{tabular}{lc|ccc|ccc|ccc|ccc}
					\toprule[1.5pt]
					\multirow{2}{4em}{\textbf{Methods}} & \multirow{2}{4em}{\textbf{Public.}} &\multicolumn{3}{c}{\textbf{TNO}} &\multicolumn{3}{c}{\textbf{RoadScene}}&\multicolumn{3}{c}{\textbf{M$^3$FD}} &\multicolumn{3}{c}{\textbf{MSIFT}} \\
					& & MSE~$\downarrow$ & NCC~$\uparrow$ & MI~$\uparrow$ & MSE~$\downarrow$ & NCC~$\uparrow$ & MI~$\uparrow$& MSE~$\downarrow$ & NCC~$\uparrow$ & MI~$\uparrow$ & MSE~$\downarrow$ & NCC~$\uparrow$ & MI~$\uparrow$\\
					\midrule[1pt]
					\specialrule{0em}{1pt}{1pt}
					Misaligned & None & $0.0067$ & $0.876$ & $1.558$  & $0.0104$ & $0.894$ & $1.602$ & $0.0054$ & $0.915$ & $2.054$ & $0.0080$ & $0.840$ & $1.408$ \\
					\cdashline{1-14}[0.8pt/2pt]
					
					\specialrule{0em}{1pt}{1pt}
					FlowNet~\cite{flownet} & CVPR'17 & $0.0292$ & $0.636$ & $1.095$ & $0.0090$ & $0.910$ & $1.549$ & 0.0357 & 0.636 & 1.638 & $\textcolor{mygreen}{\textbf{0.0014}}$ & $0.968$ & $1.894$ \\
					VoxelMorph~\cite{voxelmorph} & CVPR'18 & $0.0069$ & $0.880$ & $1.545$ & $0.0083$ & $0.914$ & $1.589$ & 0.0051 & 0.922 & 2.009 & $0.0018$ & $0.959$ & $1.776$ \\
					DGC-Net~\cite{DGCNet19} & WACV'19 & $\textcolor{mygreen}{\textbf{0.0047}}$ & $\textcolor{mygreen}{\textbf{0.928}}$ & $1.606$ & $0.0060$ & $0.939$ & $1.671$ & $0.0032$ & $0.950$ & $2.116$ & $0.0036$ & $0.923$ & $1.580$ \\
					GLU-Net~\cite{GLUNet20} & CVPR'20 & $0.0189$ & $0.719$ & $1.189$ & $0.0127 $ & $0.867$ & $1.467$ & $0.0081$ & $0.864$ & $1.798$ & $0.0053$ & $0.894$ & $1.473$ \\
					\midrule
					
					\specialrule{0em}{1pt}{1pt}
					FlowNet$^\dagger$~\cite{flownet} & CVPR'17 & $0.0073$ & $0.893$ & $1.465$ & $\textcolor{mygreen}{\textbf{0.0051}}$ & $\textcolor{mygreen}{\textbf{0.949}}$  & $\textcolor{mygreen}{\textbf{1.744}}$  & $\textcolor{mygreen}{\textbf{0.0031}}$ & $\textcolor{mygreen}{\textbf{0.951}}$  & $\textcolor{mygreen}{\textbf{2.145}}$  & $\textcolor{mygreen}{\textbf{0.0014}}$ & $\textcolor{mygreen}{\textbf{0.969}}$ & $\textcolor{myblue}{\textbf{1.899}}$ \\
					VoxelMorph$^\dagger$~\cite{voxelmorph} & CVPR'18 & $0.0050$ & $0.919$ & $1.573$ & $0.0058$ & $0.941$ & $1.689$ & 0.0033 & 0.947 & 2.103 & $0.0018$ & $0.960$ & $1.787$ \\
					DGC-Net$^\dagger$~\cite{DGCNet19} & WACV'19 & $\textcolor{myblue}{\textbf{0.0045}}$ &  $\textcolor{myblue}{\textbf{0.931}}$ & $\textcolor{mygreen}{\textbf{1.623}}$ & $0.0057$ & $0.939$ & $1.703$ & $\textcolor{mygreen}{\textbf{0.0031}}$ & $\textcolor{mygreen}{\textbf{0.951}}$ & $2.128$ & $0.0035$ & $0.926$ & $1.593$ \\
					GLU-Net$^\dagger$~\cite{GLUNet20} & CVPR'20 & $0.0197$ & $0.736$ & $1.257$ & $ 0.0112$ & $ 0.887 $ & $ 1.624 $ & $0.0085$ & $0.861$ & $1.792$ & $0.0051$ & $0.897$ & $1.481$ \\
					Nemar~\cite{Nemar_20} & CVPR'20 & $0.1401$ & $0.309$ & $0.438$ & $ 0.0814 $ & $ 0.846 $ & $ 0.986 $ & $0.1988$ & $0.033$ & $0.582$ & $0.0848$ & $0.704$ & $0.689$ \\
					CrossRAFT~\cite{CrossRAFT22} & CVPR'22 & $0.0079$ & $0.858$ & $1.602$ & $0.0088$ & $0.910$ & $\textcolor{mygreen}{\textbf{1.744}}$ & $0.0050$ & $0.917$ & $2.135$ & $0.0059$ & $0.888$ & $1.787$ \\
					SuperFusion~\cite{SuperFusion_22} & CVPR'22 & $0.0070$ & $0.886$ & $1.507$ & $0.0044$ & $0.953$ & $1.820$ & $0.0034$ & $0.942$ & $2.076$ & $0.0021$ & $0.953$ & $1.821$ \\
					\specialrule{0em}{1pt}{1pt}
					CGRP~\cite{UMF} & IJCAI'22 & $\textcolor{mygreen}{\textbf{0.0047}}$ & $0.926$ & $\textcolor{myblue}{\textbf{1.648}}$ & $\textcolor{myblue}{\textbf{0.0036}}$ & $\textcolor{myblue}{\textbf{0.963}}$  & $\textcolor{myblue}{\textbf{1.833}}$  & $\textcolor{myblue}{\textbf{0.0026}}$ & $\textcolor{myblue}{\textbf{0.957}}$  & $\textcolor{myblue}{\textbf{2.223}}$  & $\textcolor{myblue}{\textbf{0.0013}}$ & $\textcolor{myblue}{\textbf{0.970}}$  & $\textcolor{mygreen}{\textbf{1.897}}$ \\
					\midrule
					C-MPDR & None & $\textcolor{myred}{\textbf{0.0028}}$ & $\textcolor{myred}{\textbf{0.957}}$  & $\textcolor{myred}{\textbf{1.804}}$  & $\textcolor{myred}{\textbf{0.0024}}$ & $\textcolor{myred}{\textbf{0.976}}$  & $\textcolor{myred}{\textbf{1.993}}$  & $\textcolor{myred}{\textbf{0.0018}}$ & $\textcolor{myred}{\textbf{0.970}}$  & $\textcolor{myred}{\textbf{2.243}}$  & $\textcolor{myred}{\textbf{0.0009}}$ & $\textcolor{myred}{\textbf{0.980}}$  & $\textcolor{myred}{\textbf{2.028}}$  \\
					\bottomrule[1.5pt]
		\end{tabular}}		}
	\end{center}
 \vspace{-4mm}
\end{table*}

\subsection{Evaluation in IR-VIS Image Registration}
\subsubsection{\bf{Quantitative Evaluation}}
Due to the scarcity of multi-modality image registration methods, only Nemar~\cite{Nemar_20}, CrossRAFT~\cite{crossRAFT_22}, SuperFusion~\cite{SuperFusion_22}, and CGRP in UMF~\cite{UMF} are currently available. 
For a comprehensive comparison, we compare our C-MPDR with SOTA mono-modality image registration methods, including FlowNet~\cite{flownet}, VoxelMorph~\cite{voxelmorph}, DGC-Net~\cite{dgcnet_19}, and GLU-Net~\cite{glunet_20}.
In addition, we plug the CPST proposed by UMF~\cite{UMF} into the above methods to form approximate cross-modality registration variants and then conduct a relatively fair comparison. We use "$\dagger$" to mark these variants in Table~\ref{tab:align-resluts}.
By the numerical analysis, using mono-modality registration methods directly for infrared and visible images yields negligible improvement, even worse than the misaligned inputs.
Compared with cross-modality registration methods, our C-MPDR ranks first in all metrics (\ie, MSE, NCC, and MI), which outperforms other SOTA methods by large margins (\eg, +30.7\% and +8.04\% in MSE and MI metrics on M$^3$FD dataset than the latest study SuperFusion~\cite{SuperFusion_22}) on all four datasets. 
In particular, it incorporates improvements of progressive and dense deformation field estimation and demonstrates noticeable gains (\ie, +33.3\% and +8.73\% in MSE and MI on RoadScene dataset) over the CGRP in~\cite{UMF}.

\subsubsection{\bf{Qualitative Evaluation}}
\begin{figure*}[t]
	\begin{center}
		\resizebox{1.0\linewidth}{!}{
			\begin{tabular}{ccccc}	
				
				\includegraphics[width = 0.18\linewidth]{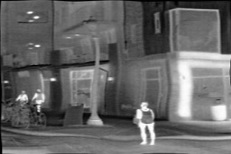} & 
				\hspace{-0.46cm}
				\includegraphics[width = 0.18\linewidth]{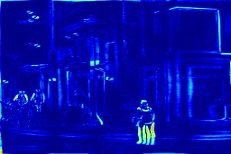}  & 
				\hspace{-0.46cm}
				\includegraphics[width = 0.18\linewidth]{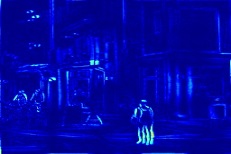}& 
				\hspace{-0.46cm}
				\includegraphics[width = 0.18\linewidth]{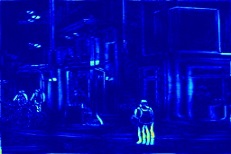} & 
				\hspace{-0.46cm}
				\includegraphics[width = 0.18\linewidth]{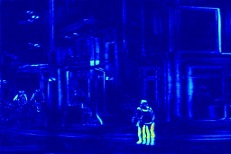} \\	
				
				\small (a) Misaligned IR
				& \hspace{-0.46cm} \small (b) IR
				& \hspace{-0.46cm} \small (c) FlowNet$^\dagger$~\cite{flownet}
				& \hspace{-0.46cm} \small (d) VM$^\dagger$~\cite{voxelmorph}
				& \hspace{-0.46cm} \small (e) DGCNet$^\dagger$~\cite{DGCNet19} \\

				\includegraphics[width = 0.18\linewidth]{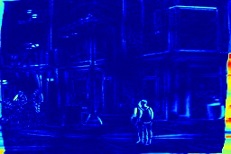}  & 
				\hspace{-0.46cm}
				\includegraphics[width = 0.18\linewidth]{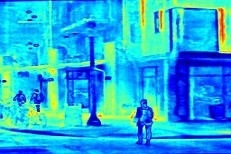} & 
				\hspace{-0.46cm}
				\includegraphics[width = 0.18\linewidth]{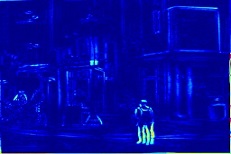} & 
				\hspace{-0.46cm}
				\includegraphics[width = 0.18\linewidth]{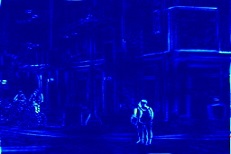}   & 
				\hspace{-0.46cm}
				\includegraphics[width = 0.18\linewidth]{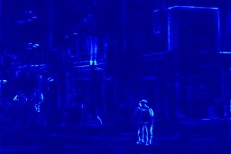}  
				\\		
				
				\small (f) GLU-Net$^\dagger$~\cite{GLUNet20}
				& \hspace{-0.46cm} \small (g) Nemar~\cite{NeMAR}
				& \hspace{-0.46cm} \small (h) CrossRAFT~\cite{CrossRAFT22}
				& \hspace{-0.46cm} \small (i) SuperFusion~\cite{SuperFusion_22}
				& \hspace{-0.46cm} \small (j) C-MPDR (Ours)
				\\
		\end{tabular}}
	\end{center}
	\vspace{-3mm}
	\caption{Error visualization of the registration results on the \textbf{RoadScene} dataset.  "$\dagger$" denotes that the CPSTN~\cite{UMF} is utilized as the basic model to mitigate cross-modality discrepancy.}
 \vspace{-2mm}
	\label{fig:reg_road}
\end{figure*}

\begin{figure*}[t]
	\begin{center}
		\resizebox{1.0\linewidth}{!}{
			\begin{tabular}{cccc}	
				
				\includegraphics[width = 0.24\linewidth, height=0.10\textheight]{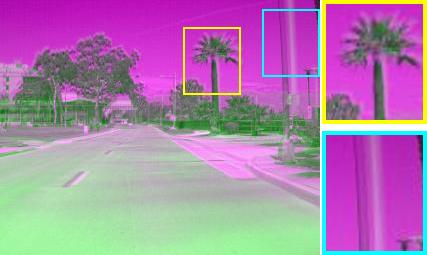}  & 
				\hspace{-0.46cm}
				\includegraphics[width = 0.24\linewidth, height=0.10\textheight]{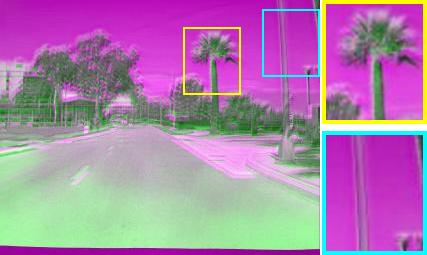}& 
				\hspace{-0.46cm}
				\includegraphics[width = 0.24\linewidth, height=0.10\textheight]{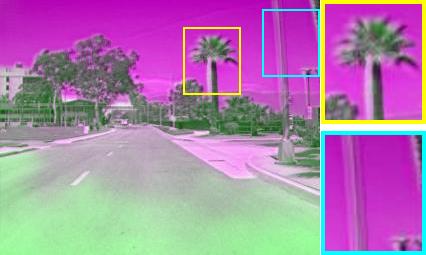} & 
				\hspace{-0.46cm}
				\includegraphics[width = 0.24\linewidth, height=0.10\textheight]{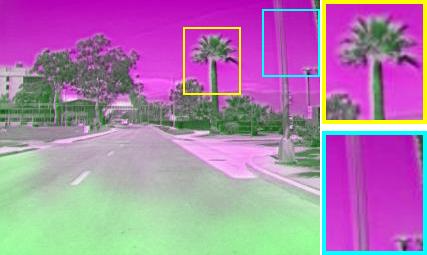} \\	
				
				\small (a) Misaligned IR
				& \hspace{-0.46cm} \small (b) FlowNet$^\dagger$~\cite{flownet}
				& \hspace{-0.46cm} \small (c) VM$^\dagger$~\cite{voxelmorph}
				& \hspace{-0.46cm} \small (d) DGCNet$^\dagger$~\cite{DGCNet19} \\

				\includegraphics[width = 0.24\linewidth, height=0.10\textheight]{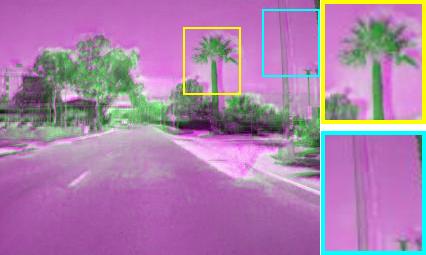}   & 
				\hspace{-0.46cm}
				\includegraphics[width = 0.24\linewidth, height=0.10\textheight]{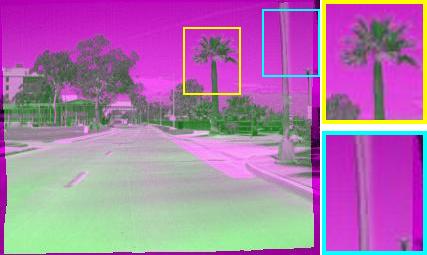} & 
				\hspace{-0.46cm}
				\includegraphics[width = 0.24\linewidth, height=0.10\textheight]{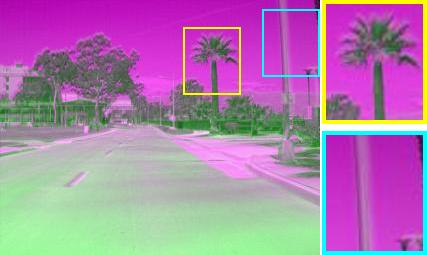} & 
				\hspace{-0.46cm}
				\includegraphics[width = 0.24\linewidth, height=0.10\textheight]{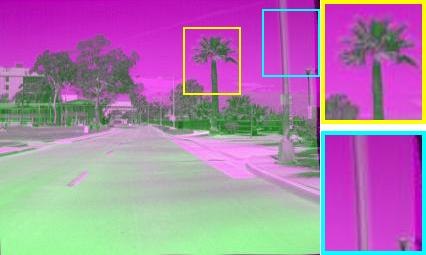}  
				\\		
				
				\small (e) Nemar~\cite{NeMAR}
				& \hspace{-0.46cm} \small (f) CrossRAFT~\cite{CrossRAFT22}
				& \hspace{-0.46cm} \small (g) SuperFusion~\cite{SuperFusion_22}
				& \hspace{-0.46cm} \small (h) C-MPDR (Ours)
				\\
		\end{tabular}}
	\end{center}
	\vspace{-3mm}
	\caption{Error visualization of the registration results on the \textbf{real-world FLIR} dataset. Noted that the CPSTN~\cite{UMF} is utilized as the basic model to mitigate cross-modality discrepancy.}
 \vspace{-4mm}
	\label{fig:reg_flir}
\end{figure*}

To intuitively exhibit the superiority of our C-MPDR, we first provide an error visualization on the synthetic misaligned RoadScene dataset in Fig.~\ref{fig:reg_road}, which shows error maps that contrast registered infrared images of different methods with the target infrared images. Brighter indicates a larger error.
It is evident that our C-MPDR demonstrates a more favorable registration performance.
Furthermore, we also provide registration cases of different methods in the real-world misaligned multi-modality dataset in Fig.~\ref{fig:reg_flir}.
As the real misaligned dataset lacks target infrared images, we evaluate the registration accuracy by the error map between the registered image and the visible image.
Upon observation, our C-MPDR produces a desirable registration effect and suppresses edge ghosts.
%
This can be attributed to two essential components within the C-MPDR subnetwork. The DFF module facilitates an effective dense representation of multi-scale deformation subfields and the PFF module achieves a progressive refinement between multi-scale registered infrared features and pseudo-infrared features.

\begin{table*}[t]
	\caption{The quantitative evaluation of the IVIF results with state-of-the-art fusion methods on three datasets using \textbf{FlowNet}~\cite{flownet} algorithm as the basic registration model. The top three results are highlighted using \textcolor{myred}{\textbf{red}}, \textcolor{myblue}{\textbf{blue}} and \textcolor{mygreen}{\textbf{green}} in the order.}
	\label{tab:flow-fus-results}
	\vspace{-6mm}
	\begin{center}
		\centering
		\setlength{\tabcolsep}{7.0pt}{
			\resizebox{1.0\linewidth}{!}{
				\begin{tabular}{l|r|c|ccccccc|cc}
					\toprule[1.5pt]
					\multirow{2}{*}{\textbf{Dataset}}  & \multirow{2}{*}{\textbf{Metric}} & FGAN & DIDFuse & PMGI & DDcGAN & RFN & MFEIF & U2F & TarDAL & CDDFuse & \multirow{2}{*}{Ours} \\
					&  & \cite{FGAN} & \cite{DIDFuse_2020} & \cite{PMGI} & \cite{DDcGAN} & \cite{RFN} & \cite{MFEIF} &  \cite{U2Fusion} & \cite{TarDAL_22} & \cite{CDDFuse23} & \\
					\midrule[1pt]
					\multirow{5}{4em}{\textit{\textbf{RoadScene}}}
					& CC~$\uparrow$   & 0.502 & 0.573 & 0.530 & 0.569 & \textcolor{mygreen}{\textbf{0.587}} & \textcolor{myblue}{\textbf{0.588}} & 0.558 & 0.544 & 0.583 & \textcolor{myred}{\textbf{0.595}} \\
					& MI~$\uparrow$ & 1.886 & 1.942 & 1.965 & 1.941 & 1.951 & 2.009 & 1.681 & \textcolor{myred}{\textbf{2.498}} & \textcolor{myblue}{\textbf{2.410}} & \textcolor{mygreen}{\textbf{2.408}} \\
					& VIF~$\uparrow$  & 0.493 & 0.445 & 0.408 & 0.538 & 0.563 & \textcolor{mygreen}{\textbf{0.619}} & 0.430 & \textcolor{myblue}{\textbf{0.710}} & 0.609 & \textcolor{myred}{\textbf{0.879}} \\
					& SSIM~$\uparrow$ & 0.255 & 0.301 & 0.266 & 0.314 & 0.319 & 0.342 & 0.297 & \textcolor{mygreen}{\textbf{0.435}} & \textcolor{myblue}{\textbf{0.468}} & \textcolor{myred}{\textbf{0.499}} \\
					& Q$_{abf}$~$\uparrow$ & 0.237 & 0.278 & 0.245 & 0.323 & 0.282 & 0.288 & 0.260 & \textcolor{mygreen}{\textbf{0.360}} & \textcolor{myblue}{\textbf{0.424}} & \textcolor{myred}{\textbf{0.427}} \\
					\midrule
					\multirow{5}{*}{\textit{\textbf{TNO}}}
					& CC~$\uparrow$   & 0.315 & 0.380 & 0.363 & 0.366 & 0.392 & 0.381 & 0.378 & \textcolor{myblue}{\textbf{0.440}} & \textcolor{mygreen}{\textbf{0.430}} & \textcolor{myred}{\textbf{0.481}} \\
					& MI~$\uparrow$ & 1.368 & 1.762 & 1.489 & 1.434 & 1.628 & 1.743 & 1.329 & \textcolor{myred}{\textbf{2.168}} & \textcolor{mygreen}{\textbf{2.024}} & \textcolor{myblue}{\textbf{2.163}} \\
					& VIF~$\uparrow$  & 0.588 & 0.545 & 0.518 & 0.510 & 0.671 & \textcolor{mygreen}{\textbf{0.701}} & 0.535 & \textcolor{myblue}{\textbf{0.715}} & 0.567 & \textcolor{myred}{\textbf{1.016}} \\
					& SSIM~$\uparrow$ & 0.236 & 0.266 & 0.277 & 0.311 & 0.300 & 0.297 & 0.299 & \textcolor{myblue}{\textbf{0.437}} & \textcolor{mygreen}{\textbf{0.338}} & \textcolor{myred}{\textbf{0.473}} \\
					& Q$_{abf}$~$\uparrow$ & 0.168 & 0.288 & 0.241 & 0.302 & 0.279 & 0.281 & 0.275 & \textcolor{myblue}{\textbf{0.389}} & \textcolor{mygreen}{\textbf{0.331}} & \textcolor{myred}{\textbf{0.427}} \\
					\midrule
					\multirow{5}{*}{\textit{\textbf{M$^3$FD}}}
					& CC~$\uparrow$ & 0.395 & 0.448 & 0.415 & 0.465 & 0.450 & 0.463 & 0.445 & \textcolor{mygreen}{\textbf{0.482}} & \textcolor{myblue}{\textbf{0.500}} & \textcolor{myred}{\textbf{0.530}} \\
					& MI~$\uparrow$ & 1.816 & 2.162 & 2.043 & 1.925 & 2.080 & 2.210 & 1.889 & \textcolor{mygreen}{\textbf{2.450}} & \textcolor{myred}{\textbf{2.796}} & \textcolor{myblue}{\textbf{2.460}} \\
					& VIF~$\uparrow$ & 0.596 & 0.555 & 0.519 & 0.451 & 0.479 & \textcolor{myblue}{\textbf{0.778}} & 0.486 & 0.643 & \textcolor{mygreen}{\textbf{0.652}} & \textcolor{myred}{\textbf{0.930}} \\
					& SSIM~$\uparrow$ & 0.330 & 0.289 & 0.319 & 0.329 & 0.277 & 0.361 & 0.314 & \textcolor{mygreen}{\textbf{0.395}} & \textcolor{myblue}{\textbf{0.442}} & \textcolor{myred}{\textbf{0.483}} \\
					& Q$_{abf}$~$\uparrow$ & 0.295 & 0.344 & 0.332 & \textcolor{mygreen}{\textbf{0.386}} & 0.292 & 0.369 & 0.349 & 0.280 & \textcolor{myred}{\textbf{0.450}} & \textcolor{myblue}{\textbf{0.417}} \\
					
					\bottomrule[1.5pt]
		\end{tabular}}}
	\end{center}
 \vspace{-4mm}
\end{table*}

\subsection{Comparisons with state-of-the-art IVIF methods}
We perform quantitative and qualitative evaluations for the proposed IMF against several state-of-the-art IVIF methods, including FGAN~\cite{FGAN}, DIDFuse~\cite{DIDFuse_2020}, PMGI~\cite{PMGI}, DDcGAN~\cite{DDcGAN}, RFN~\cite{RFN}, MFEIF~\cite{MFEIF}, U2F~\cite{U2Fusion}, TarDAL~\cite{TarDAL_22}, and CDDFuse~\cite{CDDFuse23} equipped with FlowNet~\cite{flownet} and VoxelMorph~\cite{voxelmorph} to implement cross-modality image pre-registration.
Noted that, since none of the above methods are designed for the fusion of near-infrared and visible images, we do not provide the corresponding quantitative and qualitative fusion results.

\subsubsection{\bf{Quantitative Evaluation}}
We first report the quantitative fusion results on RoadScene, TNO, and M$^3$FD datasets in Table~\ref{tab:flow-fus-results}.
According to Table~\ref{tab:align-resluts}, FlowNet$^\dagger$ shows the best registration performance except for our previous version CGRP~\cite{UMF}. To be objective, we use FlowNet$^\dagger$ as the pre-registered model of SOTA fusion methods to evaluate their quantitative results.
%
It can be seen that our IMF numerically outperforms existing IVIF methods by large margins and ranks first or second across all five metrics (\ie, CC, MI,  VIF, SSIM, and Q$_{abf}$) in these three datasets.
Especially, compared with CDDFuse~\cite{CDDFuse23} equipped with FlowNet$^\dagger$, a recent research, our IMF outperforms it overall.
The quantitative results suggest our IMF exhibits more favorable performance over other SOTA fusion methods for misaligned multi-modality images.

\subsubsection{\bf{Qualitative Evaluation}}
\begin{figure*}[t]
	\begin{center}
		\resizebox{1.0\linewidth}{!}{
			\begin{tabular}{cccccc}	
				
				\includegraphics[width = 0.146\linewidth]{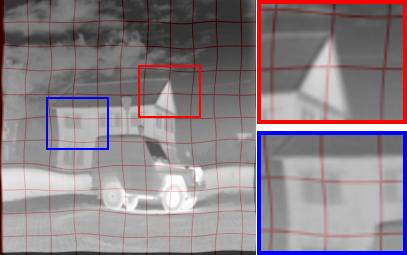} & 
				\hspace{-0.46cm}
				\includegraphics[width = 0.146\linewidth]{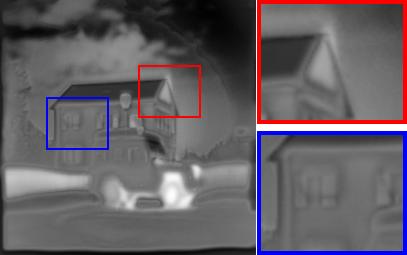}  & 
				\hspace{-0.46cm}
				\includegraphics[width = 0.146\linewidth]{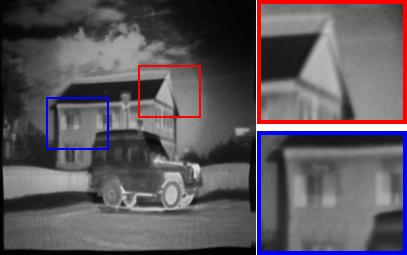}& 
				\hspace{-0.46cm}
				\includegraphics[width = 0.146\linewidth]{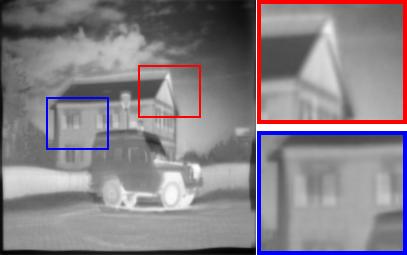} & 
				\hspace{-0.46cm}
				\includegraphics[width = 0.146\linewidth]{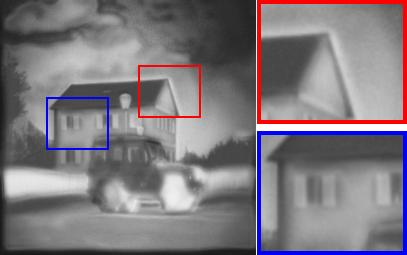} & 
				\hspace{-0.46cm}
				\includegraphics[width = 0.146\linewidth]{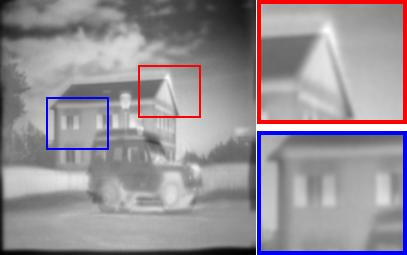} \\	
				
				\small IR
				& \hspace{-0.46cm} \small FGAN~\cite{FGAN}
				& \hspace{-0.46cm} \small DIDFuse~\cite{DIDFuse_2020}
				& \hspace{-0.46cm} \small PMGI~\cite{PMGI}
				& \hspace{-0.46cm} \small DDcGAN~\cite{DDcGAN} 
				& \hspace{-0.46cm} \small RFN~\cite{RFN} \\

				\includegraphics[width = 0.146\linewidth]{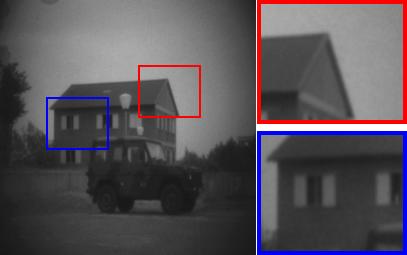}  & \hspace{-0.46cm}
				\includegraphics[width = 0.146\linewidth]{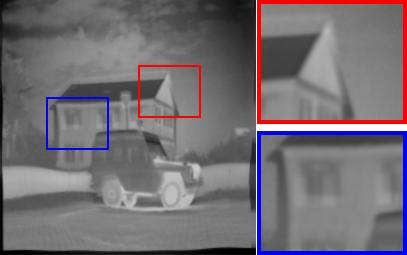}   & \hspace{-0.46cm}
				\includegraphics[width = 0.146\linewidth]{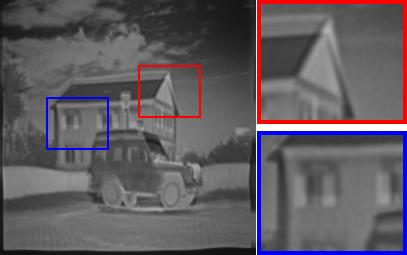} & \hspace{-0.46cm}
				\includegraphics[width = 0.146\linewidth]{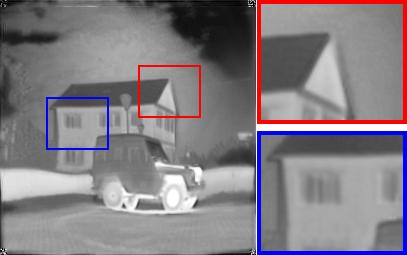}    & \hspace{-0.46cm}
				\includegraphics[width = 0.146\linewidth]{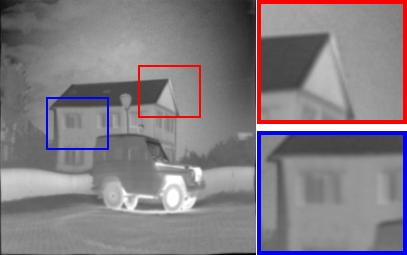} & \hspace{-0.46cm}
				\includegraphics[width = 0.146\linewidth]{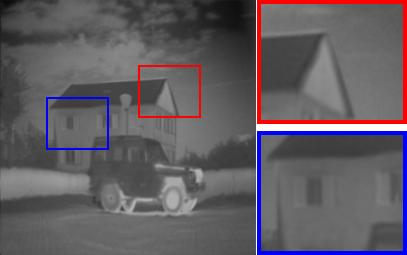}  \\		
				
				\small VIS
				& \hspace{-0.46cm} \small RFN~\cite{RFN}
				& \hspace{-0.46cm} \small U2F~\cite{U2Fusion}
				& \hspace{-0.46cm} \small TarDAL~\cite{TarDAL_22}
				& \hspace{-0.46cm} \small CDDFuse~\cite{CDDFuse23}
				& \hspace{-0.46cm} \small Ours			
				\\
		\end{tabular}}
	\end{center}
	\vspace{-3mm}
	\caption{Visualization of fusion results from the state-of-the-art infrared and visible image methods on the \textit{TNO} dataset when using the \textbf{FlowNet}~\cite{flownet} as the basic registration model. Best viewed on screen.}
 \vspace{-2mm}
	\label{fig:flownet_tno}
\end{figure*}

\begin{figure*}[t]
	\begin{center}
		\resizebox{1.0\linewidth}{!}{
			\begin{tabular}{cccccc}	
				
				\includegraphics[width = 0.146\linewidth]{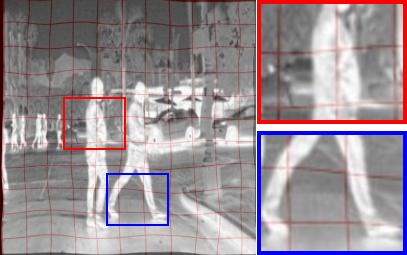} & 
				\hspace{-0.46cm}
				\includegraphics[width = 0.146\linewidth]{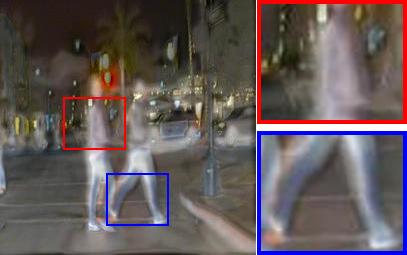}  & 
				\hspace{-0.46cm}
				\includegraphics[width = 0.146\linewidth]{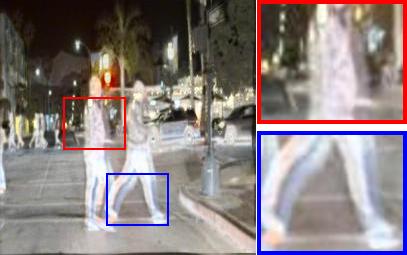}& 
				\hspace{-0.46cm}
				\includegraphics[width = 0.146\linewidth]{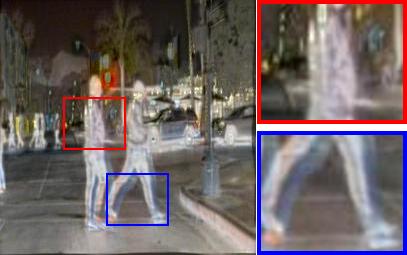} & 
				\hspace{-0.46cm}
				\includegraphics[width = 0.146\linewidth]{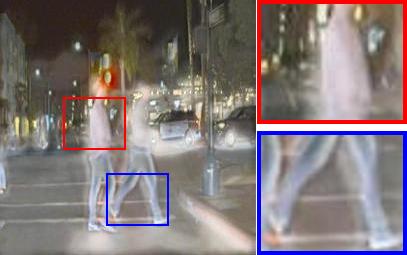} & 
				\hspace{-0.46cm}
				\includegraphics[width = 0.146\linewidth]{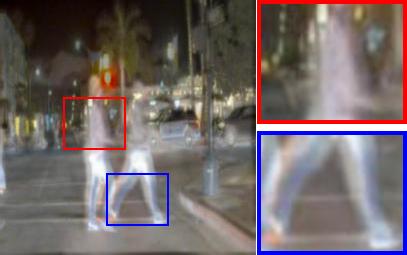} \\	
				
				\small IR
				& \hspace{-0.46cm} \small FGAN~\cite{FGAN}
				& \hspace{-0.46cm} \small DIDFuse~\cite{DIDFuse_2020}
				& \hspace{-0.46cm} \small PMGI~\cite{PMGI}
				& \hspace{-0.46cm} \small DDcGAN~\cite{DDcGAN} 
				& \hspace{-0.46cm} \small RFN~\cite{RFN} \\

				\includegraphics[width = 0.146\linewidth]{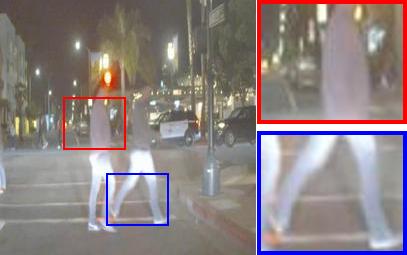}  & \hspace{-0.46cm}
				\includegraphics[width = 0.146\linewidth]{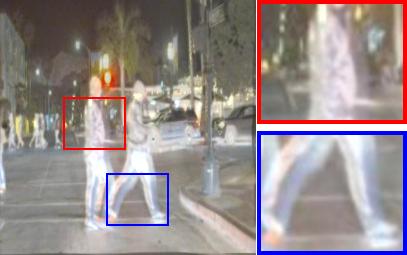}   & \hspace{-0.46cm}
				\includegraphics[width = 0.146\linewidth]{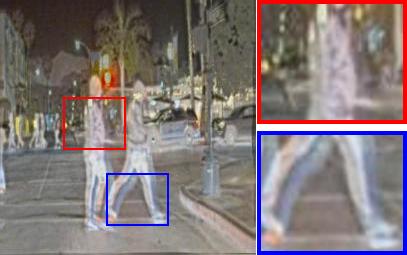} & \hspace{-0.46cm}
				\includegraphics[width = 0.146\linewidth]{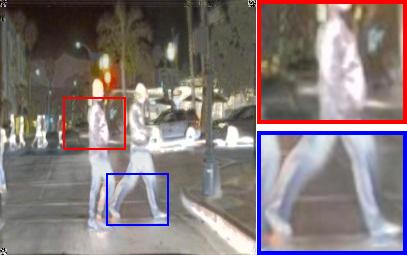}    & \hspace{-0.46cm}
				\includegraphics[width = 0.146\linewidth]{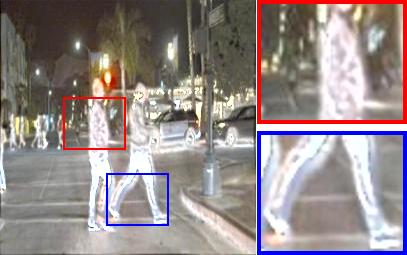} & \hspace{-0.46cm}
				\includegraphics[width = 0.146\linewidth]{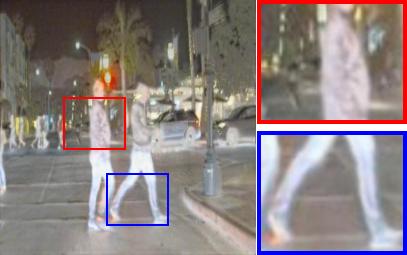} \\		
				
				\small VIS
				& \hspace{-0.46cm} \small MFEIF~\cite{MFEIF}
				& \hspace{-0.46cm} \small U2F~\cite{U2Fusion}
				& \hspace{-0.46cm} \small TarDAL~\cite{TarDAL_22}
				& \hspace{-0.46cm} \small CDDFuse~\cite{CDDFuse23}
				& \hspace{-0.46cm} \small Ours			
				\\
		\end{tabular}}
	\end{center}
	\vspace{-3mm}
	\caption{Visualization of fusion results from the state-of-the-art infrared and visible image methods on the \textit{RoadScene} dataset when using the \textbf{FlowNet}~\cite{flownet} as the basic registration model. Best viewed on screen.}
 \vspace{-4mm}
	\label{fig:flownet_road}
\end{figure*}

\begin{figure*}[t]
	\begin{center}
		\resizebox{1.0\linewidth}{!}{
			\begin{tabular}{cccccc}	
				
				\includegraphics[width = 0.146\linewidth]{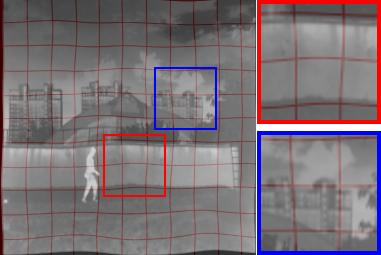} & 
				\hspace{-0.46cm}
				\includegraphics[width = 0.146\linewidth]{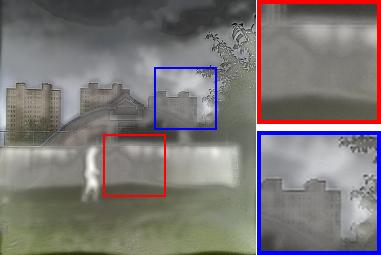}  & 
				\hspace{-0.46cm}
				\includegraphics[width = 0.146\linewidth]{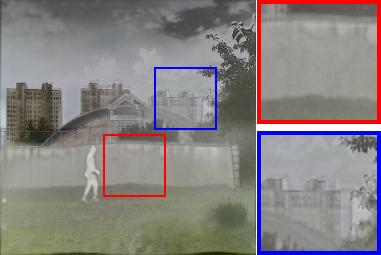} & 
				\hspace{-0.46cm}
				\includegraphics[width = 0.146\linewidth]{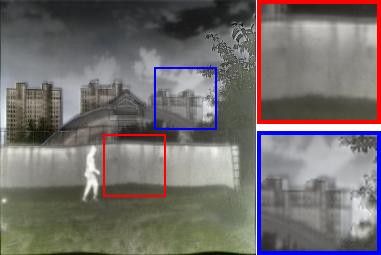} & 
				\hspace{-0.46cm}
				\includegraphics[width = 0.146\linewidth]{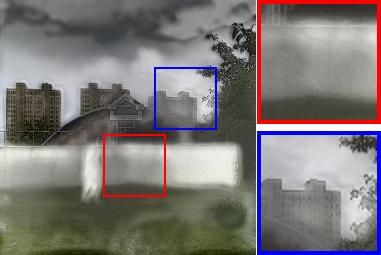} & 
				\hspace{-0.46cm}
				\includegraphics[width = 0.146\linewidth]{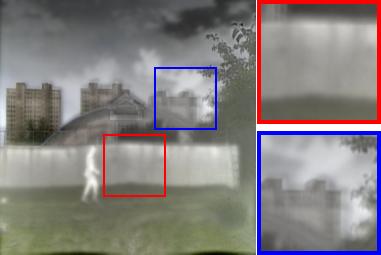}
				\\	
				
				\small IR
				& \hspace{-0.46cm} \small FGAN~\cite{FGAN}
				& \hspace{-0.46cm} \small DIDFuse~\cite{DIDFuse_2020}
				& \hspace{-0.46cm} \small PMGI~\cite{PMGI}
				& \hspace{-0.46cm} \small DDcGAN~\cite{DDcGAN} 
				& \hspace{-0.46cm} \small RFN~\cite{RFN} \\

				\includegraphics[width = 0.146\linewidth]{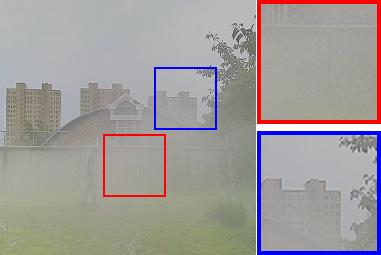} & \hspace{-0.46cm}
				\includegraphics[width = 0.146\linewidth]{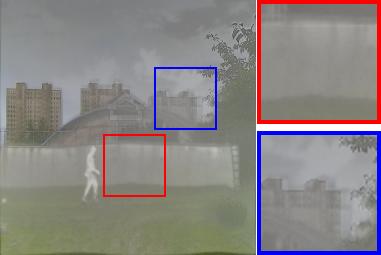}& \hspace{-0.46cm}
				\includegraphics[width = 0.146\linewidth]{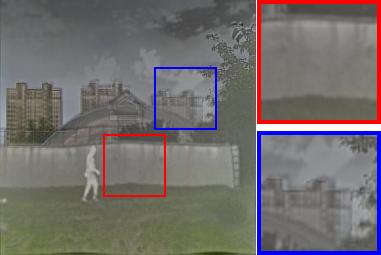} & \hspace{-0.46cm}
				\includegraphics[width = 0.146\linewidth]{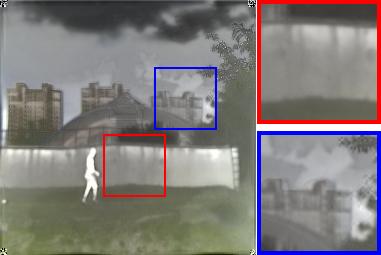} & \hspace{-0.46cm}
				\includegraphics[width = 0.146\linewidth]{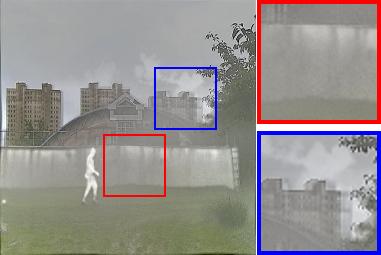}& \hspace{-0.46cm}
				\includegraphics[width = 0.146\linewidth]{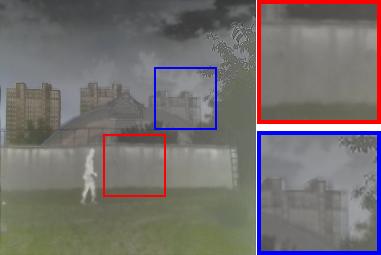} \\		
				
				\small VIS
				& \hspace{-0.46cm} \small MFEIF~\cite{MFEIF}
				& \hspace{-0.46cm} \small U2F~\cite{U2Fusion}
				& \hspace{-0.46cm} \small TarDAL~\cite{TarDAL_22}
				& \hspace{-0.46cm} \small CDDFuse~\cite{CDDFuse23}
				& \hspace{-0.46cm} \small Ours			
				\\
		\end{tabular}}
	\end{center}
 \vspace{-4mm}
	\caption{Visualization of fusion results from the state-of-the-art infrared and visible image methods on the \textit{M$^3$FD} dataset when using the \textbf{FlowNet}~\cite{flownet} as the basic registration model. Best viewed on screen.}
 \vspace{-3mm}
	\label{fig:flownet_m3fd}
\end{figure*}

\begin{figure*}[t]
	\begin{center}
		\resizebox{1.0\linewidth}{!}{
			\begin{tabular}{cccccc}	
				
				\includegraphics[width = 0.146\linewidth]{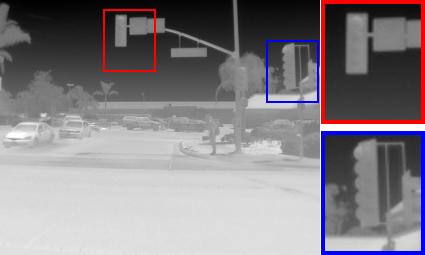}    & 
				\hspace{-0.46cm}
				\includegraphics[width = 0.146\linewidth]{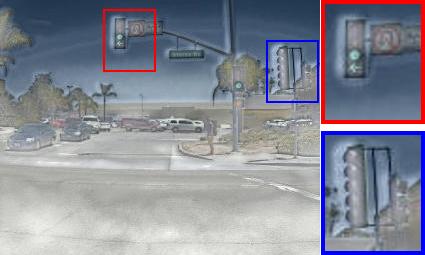}& 
				\hspace{-0.46cm}
				\includegraphics[width = 0.146\linewidth]{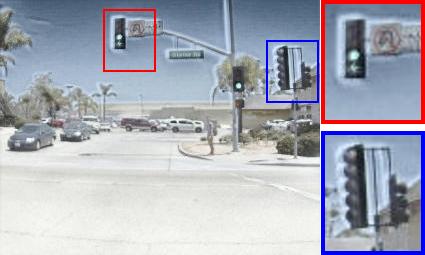} & 
				\hspace{-0.46cm}
				\includegraphics[width = 0.146\linewidth]{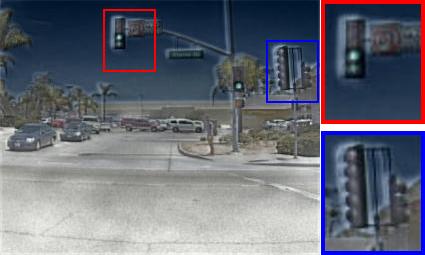}   & 
				\hspace{-0.46cm}
				\includegraphics[width = 0.146\linewidth]{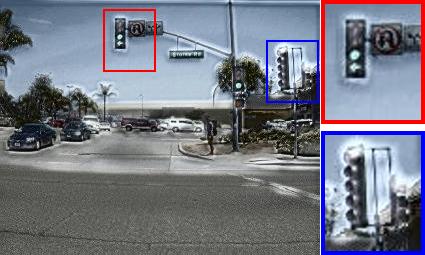} & 
				\hspace{-0.46cm}
				\includegraphics[width = 0.146\linewidth]{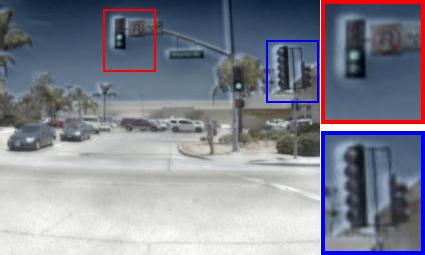}
				\\	
				
				\small IR
				& \hspace{-0.46cm} \small FGAN~\cite{FGAN}
				& \hspace{-0.46cm} \small DIDFuse~\cite{DIDFuse_2020}
				& \hspace{-0.46cm} \small PMGI~\cite{PMGI}
				& \hspace{-0.46cm} \small DDcGAN~\cite{DDcGAN} 
				& \hspace{-0.46cm} \small RFN~\cite{RFN} \\

				\includegraphics[width = 0.146\linewidth]{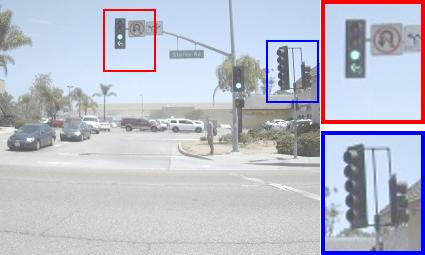}      & \hspace{-0.46cm}
				\includegraphics[width = 0.146\linewidth]{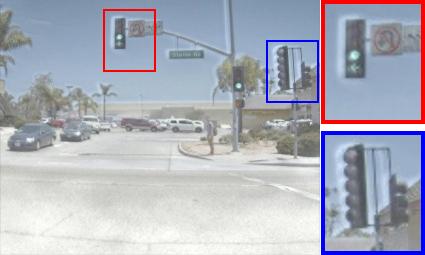}    & \hspace{-0.46cm}
				\includegraphics[width = 0.146\linewidth]{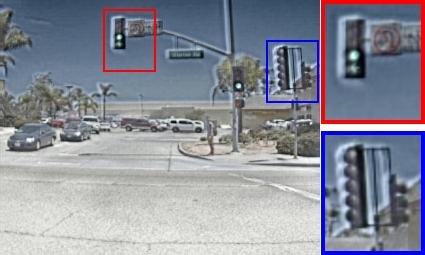} & \hspace{-0.46cm}
				\includegraphics[width = 0.146\linewidth]{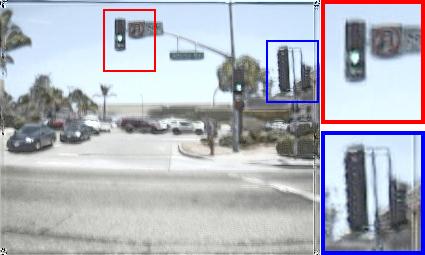}   & \hspace{-0.46cm}
				\includegraphics[width = 0.146\linewidth]{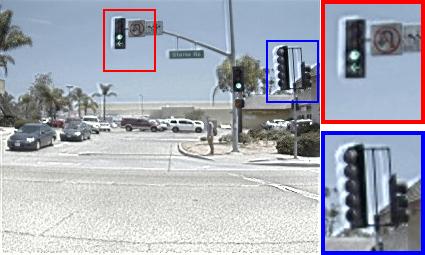}  & \hspace{-0.46cm}
				\includegraphics[width = 0.146\linewidth]{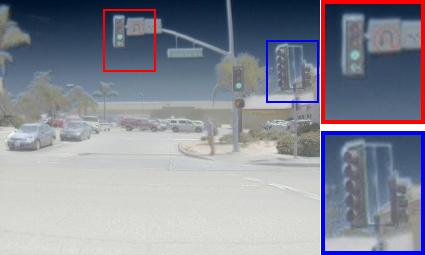} \\		
				
				\small VIS
				& \hspace{-0.46cm} \small MFEIF~\cite{MFEIF}
				& \hspace{-0.46cm} \small U2F~\cite{U2Fusion}
				& \hspace{-0.46cm} \small TarDAL~\cite{TarDAL_22}
				& \hspace{-0.46cm} \small CDDFuse~\cite{CDDFuse23}
				& \hspace{-0.46cm} \small Ours			
				\\
		\end{tabular}}
	\end{center}
 \vspace{-4mm}
	\caption{Visualization of fusion results from the state-of-the-art infrared and visible image methods on the real-world misaligned \textit{FLIR} dataset when using the \textbf{FlowNet}~\cite{flownet} as the basic registration model. Best viewed on screen.}
 \vspace{-3mm}
	\label{fig:flownet_FLIR}
\end{figure*}

We show qualitative results on three synthetic misaligned multi-modality datasets (\ie, TNO, RoadScene, and M$^3$FD) in Fig.~\ref{fig:flownet_tno}, Fig.~\ref{fig:flownet_road}, and Fig.~\ref{fig:flownet_m3fd}, respectively.
According to Table~\ref{tab:align-resluts}, FlowNet$^\dagger$ shows the best registration performance except for our previous version CGRP~\cite{UMF}. Therefore, the qualitative fusion results of the SOTA methods are based on the FlowNet$^\dagger$ registration model.
Examining zoomed-in regions reveals that the prior works result in edge ghosts due to poor registration. While our IMF corrects structural distortions, suppresses edge ghosts, and preserves sharp structures and rich textures in fused images.
In addition, we visually compare fusion results on a real-world misaligned infrared-visible image dataset (\ie, FLIR) in Fig.~\ref{fig:flownet_FLIR} using FlowNet$^\dagger$ pre-registered model. 
As can be seen, the fused image by our IMF exhibits negligible edge ghosts compared to others.
It is indicated that our C-MPDR significantly improves the quality of image fusion. Additionally, the proposed IMF framework also be demonstrated to be effective in coping with imperfect misaligned data.
\vspace{-3mm}

\subsection{Ablation studies}
\begin{table}[t]
	\scriptsize
	\caption{Quantitative ablation studies of the C-MPDR network on TNO, RoadScene, and M$^3$FD datasets.}
        \vspace{-5mm}
	\label{tab:our_ablation}
	\begin{center}
		
		\renewcommand\arraystretch{1.2} 
		\setlength{\tabcolsep}{4.0pt}{ 
			\begin{tabular}{cccccc}
				\toprule[1.2pt]
				\multirow{2}{3.3em}{\textbf{\textit{Dataset}}}&\multirow{2}{4.3em}{\textbf{C-MPDR}}&\multirow{2}{3.3em}{\textbf{TCF}}&\multicolumn{3}{c}{\textbf{\textit{Metrics}}} \\
				& & &CC$\uparrow$ & SSIM$\uparrow$ & VIF$\uparrow$\\
				\midrule
				\specialrule{0em}{1pt}{1pt}
				\multirow{2}{3.3em}{\textit{\textbf{TNO}}}&\xmark & \cmark & $0.438~~~~~~~~$ & $0.369~~~~~~~~$ & $0.732~~~~~~~~$ \\

				\specialrule{0em}{1pt}{1pt}
				&\cmark & \cmark & $\textbf{0.481}_{\color{red}{+{0.044}}}$ & $\textbf{0.473}_{\color{red}{+{0.103}}}$ & $\textbf{1.016}_{\color{red}{+{0.374}}}$ \\
				\midrule
				\specialrule{0em}{1pt}{1pt}
				\multirow{2}{3.3em}{\textit{\textbf{Road}}}&\xmark & \cmark & $0.563~~~~~~~~$ & $0.367~~~~~~~~$ & $0.543~~~~~~~~$ \\

				\specialrule{0em}{1pt}{1pt}
				&\cmark & \cmark & $\textbf{0.595}_{\color{red}{+{0.032}}}$ & $\textbf{0.499}_{\color{red}{+{0.132}}}$ & $\textbf{0.879}_{\color{red}{+{0.336}}}$ \\
				
				\midrule
				\specialrule{0em}{1pt}{1pt}
				\multirow{2}{3.3em}{\textit{\textbf{M$^3$FD}}} & \xmark & \cmark & $0.493~~~~~~~~$ & $0.348~~~~~~~~$ & $0.663~~~~~~~~$ \\
				\specialrule{0em}{1pt}{1pt}
				&\cmark & \cmark & $\textbf{0.530}_{\color{red}{+{0.037}}}$ & $\textbf{0.483}_{\color{red}{+{0.135}}}$ & $\textbf{0.930}_{\color{red}{+{0.276}}}$ \\				
				\bottomrule[1.2pt]
		\end{tabular}}
	\end{center}
 \vspace{-6mm}
\end{table}

\subsubsection{\bf{\textit{Effectiveness of C-MPDR}}}
\begin{figure}[t]
	\begin{center}
		\begin{tabular}{cc}
			\includegraphics[width = 0.435\linewidth ,height=0.12\textheight]{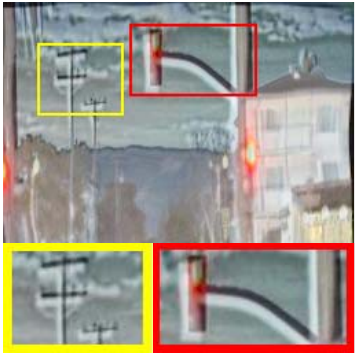} & \hspace{-0.36cm}
			\includegraphics[width = 0.435\linewidth,height=0.12\textheight]{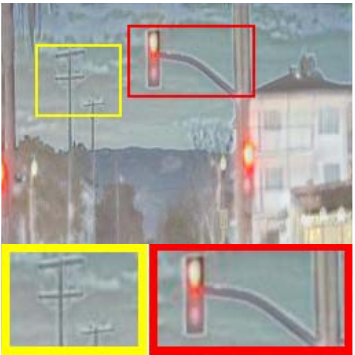} \\						
			
			{\small (a) w/o $\mathcal{R}$}
			& \hspace{-0.36cm} {\small (b) w/ $\mathcal{R}$ (Ours)}\\				
			
		\end{tabular}
	\end{center}
	\vspace{-4mm}
	\caption{Ablation analysis of the MPDR ($\mathcal{R}$) on the RoadScene dataset.}
        \vspace{-3mm}
	\label{fig:our_ablation}
\end{figure}

To investigate the effectiveness of our C-MPDR, we disable it and solely employ the fusion subnetwork TCF, utilizing paired infrared and visible images as input. 
Comparing the quantitative results on TNO, RoadScene, and M$^3$FD datasets in Table~\ref{tab:our_ablation}, it is evident that our C-MPDR improves the performance of the misaligned IVIF task to some degree, as demonstrated by gains of $0.374$, $0.336$, and $0.276$ in the VIF metric, as well as gains of $0.103$, $0.132$, and $0.135$ in the SSIM metric across the three datasets, respectively. 
A visualization example on the RoadScene dataset is shown in Fig.~\ref{fig:our_ablation}, which reveals a similar trend to that in Table.~\ref{tab:our_ablation},~\ie, the C-MPDR effectively facilitates the registration of misaligned infrared-visible images.

\begin{table}[t]
	\scriptsize
	\caption{Ablation studies of the DFF and PFF modules on the RoadScene dataset.}
        \vspace{-7mm}
	\label{tab:dff_pff_ablation}
	\begin{center}
		
		\renewcommand\arraystretch{1.1} 
		\setlength{\tabcolsep}{6.0pt}{ 
			\resizebox{1.0\linewidth}{!}{
			\begin{tabular}{ccc|cccc}
				\toprule[1.2pt]
				\textbf{No.}&\textbf{DFF} & \textbf{PFF} & MSE$\downarrow$ & NCC$\uparrow$ & MI$\uparrow$\\
				\midrule
				\specialrule{0em}{1pt}{1pt}
				1 & Only Interpolation & Concatenation & $0.0043$ & $0.956$ & $1.804$ \\
				2 & Only Interpolation & \cmark & $0.0042$ & $0.958$ & $1.817$ \\
				3 & \cmark & Concatenation & $0.0041$ & $0.958$ & $1.818$ \\
				\specialrule{0em}{1pt}{1pt}
				4 & \cmark & \cmark & $\textbf{0.0024}$ & $\textbf{0.967}$ & $\textbf{1.993}$ \\	
				\bottomrule[1.2pt]
		\end{tabular}}}
	\end{center}
 \vspace{-8mm}
\end{table}

\begin{table}[t]
	\scriptsize
	\caption{Ablation studies of the TCB and DAU modules on the RoadScene dataset.}
        \vspace{-3mm}
	\label{tab:TCB_DAU_ablation}
	\begin{center}
		\vspace{-4mm}
		\renewcommand\arraystretch{1.2} 
		\setlength{\tabcolsep}{5.8pt}{ 
			\resizebox{1.0\linewidth}{!}{
			\begin{tabular}{ccc|cccccc}
				\toprule[1.2pt]
				\textbf{No.}&\textbf{TCB} & \textbf{DAU} & CC$\uparrow$ & SSIM$\uparrow$ & VIF$\uparrow$ & MI$\uparrow$ & Q$_{abf}$$\uparrow$ \\
				\midrule
				\specialrule{0em}{1pt}{1pt}
				1 & DenseBlock & \xmark & $0.592$ & $0.343$ & $0.600$ & $1.950$ & $0.233$ \\
				2 & \cmark & \xmark & $0.593$ & $0.352$ & $0.605$ & $1.968$ & $0.258$  \\
				3 & DenseBlock & \cmark & $\textbf{0.596}$ & $0.350$ & $0.606$ & $1.961$ & $0.248$ \\
				\specialrule{0em}{1pt}{1pt}
				4 & \cmark & \cmark & $0.595$ & $\textbf{0.499}$ & $\textbf{0.879}$ & $\textbf{2.408}$ & $\textbf{0.427}$ \\				
				\bottomrule[1.2pt]
		\end{tabular}}}
	\end{center}
 \vspace{-6mm}
\end{table}

\subsubsection{\bf{\textit{Effectiveness of DFF and PFF modules}}}
\begin{figure}[t]
	\begin{center}
		\begin{tabular}{cc}
			\includegraphics[width = 0.43\linewidth, height=0.09\textheight]{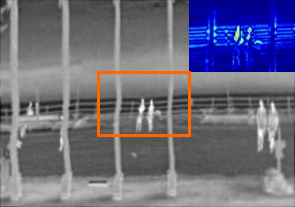} & \hspace{-0.36cm}
			\includegraphics[width = 0.43\linewidth,height=0.09\textheight]{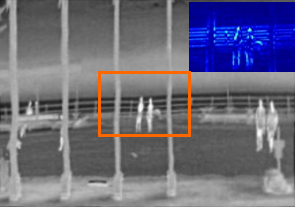} \\	
			(a) w/o DFF, PFF & \hspace{-0.36cm} 
			(b) w/o DFF, w/ PFF
			\\
			\includegraphics[width = 0.43\linewidth,height=0.09\textheight]{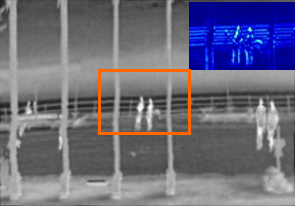} & \hspace{-0.36cm}
			\includegraphics[width = 0.43\linewidth,height=0.09\textheight]{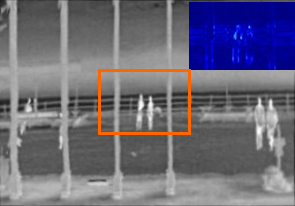} \\		
			(c) w/ DFF, w/o PFF & \hspace{-0.36cm} 
			(d) w/ DFF, PFF		
		\end{tabular}
	\end{center}
	\vspace{-3mm}
	\caption{Qualitative ablation analysis of the DFF and PFF modules from the registration perspective on the RoadScene dataset.}
        \vspace{-4mm}
	\label{fig:ab_dff_pff}
\end{figure}

We investigate the effectiveness of DFF and PFF modules by disabling their core components (\ie, the reweighting of multi-scale deformation fields in DFF and the attention-guided feature re-response in PFF) and replace them using ``Only Interpolation" and ``Concatenation" operations.
As reported in Table~\ref{tab:dff_pff_ablation}, disabling both the DFF and PFF modules results in a large degradation of the registration performance (\ie, -44.2\% in MSE, -0.011 and -0.189 in NCC and MI metrics).
And, disabling either DFF or PFF resulted in a significant decrease (\ie, -0.176 and -0.175) in the MI metric.
In addition, Fig.~\ref{fig:ab_dff_pff} shows qualitative results and local error maps. ``No.1$-$4" in Table~\ref{tab:dff_pff_ablation} corresponds to Fig.~\ref{fig:ab_dff_pff}(a)$-$(d). 
By observation, we can find that disabling either DFF or PFF leads to remaining structural distortions that are irreparable.
The quantitative and qualitative results indicate that the proposed DFF and PFF modules make a significant contribution to the more accurate estimation of deformation fields.

\subsubsection{\bf{\textit{Effectiveness of TCB and DAU modules}}}
\begin{figure*}[h]
	\begin{center}
		\resizebox{1.0\linewidth}{!}{
			\begin{tabular}{ccccc}	
				
				\includegraphics[width = 0.19\linewidth]{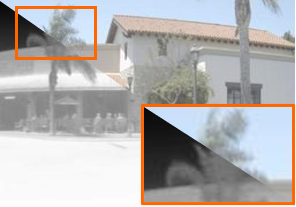}    & 
				\hspace{-0.46cm}
				\includegraphics[width = 0.19\linewidth]{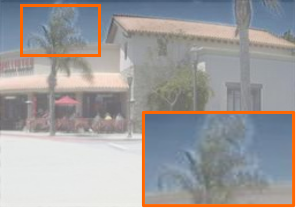} & 
				\hspace{-0.46cm}
				\includegraphics[width = 0.19\linewidth]{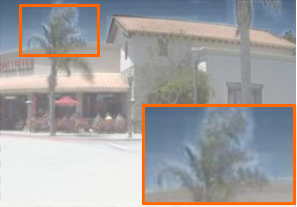}   & 
				\hspace{-0.46cm}
				\includegraphics[width = 0.19\linewidth]{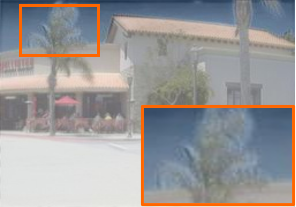} & 
				\hspace{-0.46cm}
				\includegraphics[width = 0.19\linewidth]{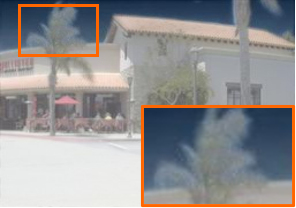}
				\\	
				
				\small (a) IR/VIS
				& \hspace{-0.46cm} \small (b) w/o TCB, DAU
				& \hspace{-0.46cm} \small (c) w/ TCB, w/o DAU
				& \hspace{-0.46cm} \small (d) w/o TCB, w/ DAU
				& \hspace{-0.46cm} \small (e) Ours \\
		\end{tabular}}
	\end{center}
	\vspace{-3mm}
	\caption{Qualitative ablation analysis of the TCB and DAU modules from the registration perspective on the RoadScene dataset. Best viewed on screen.}
 \vspace{-3mm}
	\label{fig:ab_TCB_DAU}
\end{figure*}

To evaluate the effectiveness of the two modules, TCB and DAU, we disable sequentially each of them to assess their individual influence.
Note that we replace the two TCB modules in our TCFN with the DenseBlock~\cite{DenseNet17} to perform the ablation study instead of removing them directly.
Comparing ``No.1\&No.2" and ``No.3\&No.4" experiments in Table~\ref{tab:TCB_DAU_ablation}, the TCB achieves better fusion results than DenseBlock mathematically.
Meanwhile, comparing ``No.2\&No.4" experiments, our model gains $0.440$, $0.274$, and $0.147$ improvements in MI, VIF, and SSIM metrics, respectively. 
There are two reasons for this: 
1) the TCB allows for modeling of local and global feature dependencies, capturing a wealth of multi-modality features,
while 2) the DAU as the fusion strategy enables adaptive selection of discriminative features for fusion.
Furthermore, Fig.~\ref{fig:ab_TCB_DAU} also supports this argument. Especially, our method outperforms others in suppressing edge ghosts and improving the fusion quality, as evidenced by the locally enlarged regions.

\subsubsection{\bf{\textit{One-stage training vs. Joint training}}}
\begin{figure}[t]
	\begin{center}
		\begin{tabular}{cc}
			\hspace{-6mm}
			\includegraphics[width = 0.48\linewidth ,height=0.129\textheight]{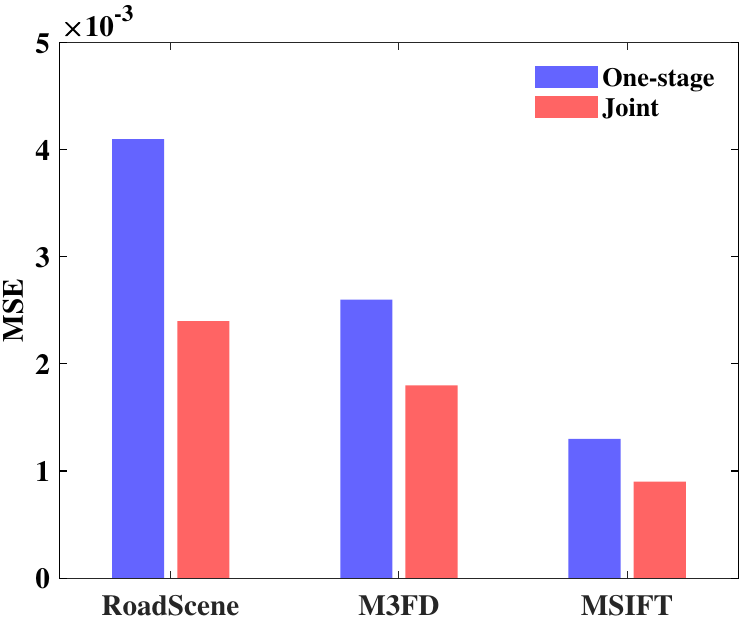} & \hspace{-0.36cm}
			\includegraphics[width = 0.49\linewidth,height=0.125\textheight]{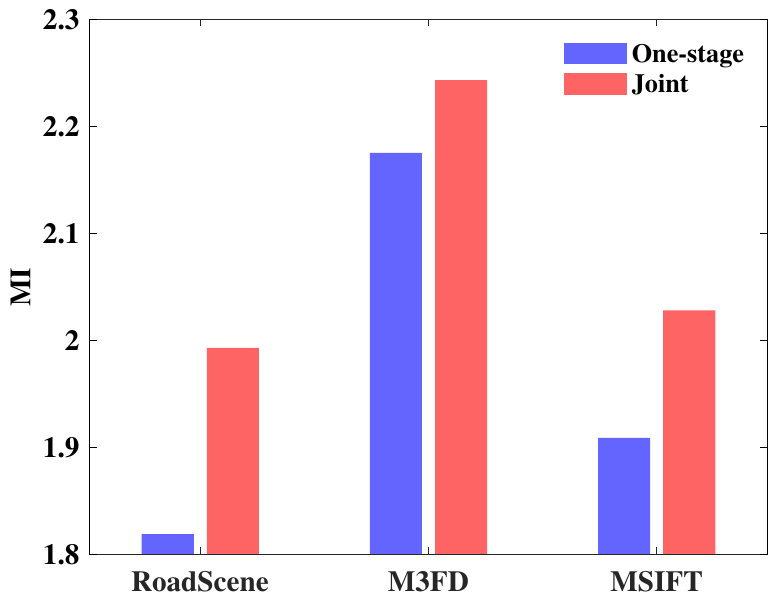} \\						
			
			\small (a) MSE 
			& \hspace{-0.36cm} \small (b) MI \\				
		\end{tabular}
	\end{center}
	\vspace{-3mm}
	\caption{Investigation on training strategies of our MPDRN and TCFN on RoadScene, M$^3$FD, and MSIFT datasets.}
 \vspace{-4mm}
	\label{fig:ab_onestage_joint}
\end{figure}

To investigate the potential of the joint training for MPDR and TCF in improving registration performance, we evaluate the registration performance of our framework with both one-stage and joint training strategies on RoadScene, M$^3$FD, and MSIFT datasets, as shown in Fig.~\ref{fig:ab_onestage_joint}. 
In one-stage training, the parameters of our MPDRN are frozen while the total fusion loss (Eq.~(\ref{eq:total_fusion_loss})) only updates the parameters of the TCFN.
Conversely, joint training involves MPDRN updating its parameters under the joint constraint of registration loss (Eq.~(\ref{eq:total_reg_loss})) and fusion loss.
Fig.~\ref{fig:ab_onestage_joint} shows the performance difference of models trained by different training strategies on the two metrics \ie, MSE and MI. Lower MSE is better and larger MI is better.
According to Fig.~\ref{fig:ab_onestage_joint}, joint training of MPDRN and TCFN improves registration performance. Considering that higher registration accuracy contributes to a better fusion image, we can conclude that image fusion and registration reinforce each other.

\subsection{Robustness Analysis}
\begin{figure}[t]
	\begin{center}
		\begin{tabular}{ccc}
			\includegraphics[width = 0.31\linewidth]{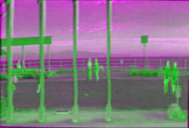}     & \hspace{-0.46cm}
			\includegraphics[width = 0.31\linewidth]{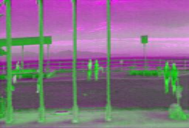} & \hspace{-0.46cm}
			\includegraphics[width = 0.31\linewidth]{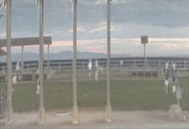} \\	
			(a) Slight & \hspace{-0.46cm} 
			(b) Registered-L & \hspace{-0.46cm} 
			(c) Fused-L \\
			
			\includegraphics[width = 0.31\linewidth]{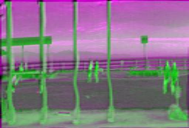}     & \hspace{-0.46cm}
			\includegraphics[width = 0.31\linewidth]{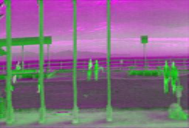} & \hspace{-0.46cm} 
			\includegraphics[width = 0.31\linewidth]{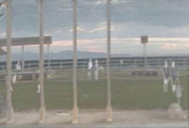} \\		
			(d) Moderate & \hspace{-0.46cm} 
			(e) Registered-M & \hspace{-0.46cm} 
			(f)	Fused-M \\
			
			\includegraphics[width = 0.31\linewidth]{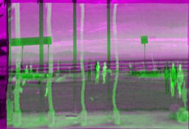}     & \hspace{-0.46cm}
			\includegraphics[width = 0.31\linewidth]{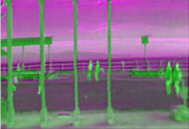} & \hspace{-0.46cm}
			\includegraphics[width = 0.31\linewidth]{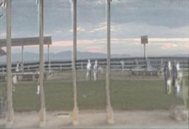} \\		
			(g) Severe & \hspace{-0.46cm} 
			(h) Registered-S & \hspace{-0.46cm}
			(i) Fused-S  	
		\end{tabular}
	\end{center}
	\vspace{-3mm}
	\caption{Visualization of registration and fusion results of infrared-visible image pairs with varying degrees of deformation on the RoadScene dataset.}
 \vspace{-3mm}
	\label{fig:slight_moderate_severe}
\end{figure}

We examine the robustness of our proposed IMF by using it for joint registration and fusion of misaligned multi-modality images with different levels of deformation, as described in Table~\ref{tab:deformable_parameters}.
Specifically, we simulate three groups of infrared and visible images with slight (\textbf{L}), moderate (\textbf{M}), and severe (\textbf{S}) deformation to test the pretrained model, thus obtaining the registered-infrared images and final fusion images.
As shown in Fig.~\ref{fig:slight_moderate_severe}, the first two columns show the error maps between the source distorted- or registered-infrared images and visible images, providing an intuitive display of the offset between input image pairs and the accuracy of registration results. The last column shows fusion images.
These results indicate that our IMF can effectively cope with different degrees of deformation, not only correcting structural distortions and eliminating offset but also suppressing edge ghosts in the fusion images, thus significantly improving image quality.

\section{Conclusion}\label{conclusion}
In this work, we proposed an improving fusion framework called IMF to cope with the challenges brought by the misalignments between infrared and visible sensors.
On the one hand, this framework emphasizes the customization of a one-stage progressive dense multi-modality image registration method, with the aim of providing well-aligned images to support the oriented image fusion task.
On the other hand, this framework involves a Transformer-Conv-based fusion network equipped with a dual-attention fusion strategy, not only modeling local and global multi-modality feature representations but also adaptively selecting salient and meaningful features for fusion.
In the end, the proposed framework successfully mitigated the structural distortions and edge ghosts, resulting in high-quality fused images.


%





\ifCLASSOPTIONcaptionsoff
  \newpage
\fi



\bibliographystyle{IEEEtran}
\bibliography{Transactions-Bibliography/IEEEfull}

\begin{thebibliography}{10}
\providecommand{\url}[1]{#1}
\csname url@samestyle\endcsname
\providecommand{\newblock}{\relax}
\providecommand{\bibinfo}[2]{#2}
\providecommand{\BIBentrySTDinterwordspacing}{\spaceskip=0pt\relax}
\providecommand{\BIBentryALTinterwordstretchfactor}{4}
\providecommand{\BIBentryALTinterwordspacing}{\spaceskip=\fontdimen2\font plus
\BIBentryALTinterwordstretchfactor\fontdimen3\font minus
  \fontdimen4\font\relax}
\providecommand{\BIBforeignlanguage}[2]{{%
\expandafter\ifx\csname l@#1\endcsname\relax
\typeout{** WARNING: IEEEtran.bst: No hyphenation pattern has been}%
\typeout{** loaded for the language `#1'. Using the pattern for}%
\typeout{** the default language instead.}%
\else
\language=\csname l@#1\endcsname
\fi
#2}}
\providecommand{\BIBdecl}{\relax}
\BIBdecl

\bibitem{CDDFuse23}
Z.~Zhao, H.~Bai, J.~Zhang, Y.~Zhang, S.~Xu, Z.~Lin, R.~Timofte, and
  L.~Van~Gool, ``Cddfuse: Correlation-driven dual-branch feature decomposition
  for multi-modality image fusion,'' in \emph{CVPR}, 2023.

\bibitem{TarDAL_22}
J.~Liu, X.~Fan, Z.~Huang, G.~Wu, R.~Liu, W.~Zhong, and Z.~Luo, ``Target-aware
  dual adversarial learning and a multi-scenario multi-modality benchmark to
  fuse infrared and visible for object detection,'' in \emph{CVPR}, 2022.

\bibitem{IRFS_23}
D.~Wang, J.~Liu, R.~Liu, and X.~Fan, ``An interactively reinforced paradigm for
  joint infrared-visible image fusion and saliency object detection,''
  \emph{Information Fusion}, vol.~98, p. 101828, 2023.

\bibitem{FGAN}
J.~Ma, W.~Yu, P.~Liang, C.~Li, and J.~Jiang, ``Fusiongan: {A} generative
  adversarial network for infrared and visible image fusion,''
  \emph{Information Fusion}, vol.~48, pp. 11--26, 2019.

\bibitem{DIDFuse_2020}
Z.~Zhao, S.~Xu, C.~Zhang, J.~Liu, J.~Zhang, and P.~Li, ``Didfuse: Deep image
  decomposition for infrared and visible image fusion,'' in \emph{IJCAI}, 2020,
  pp. 970--976.

\bibitem{PMGI}
H.~Zhang, H.~Xu, Y.~Xiao, X.~Guo, and J.~Ma, ``Rethinking the image fusion: {A}
  fast unified image fusion network based on proportional maintenance of
  gradient and intensity,'' in \emph{AAAI}, 2020, pp. 12\,797--12\,804.

\bibitem{DDcGAN}
J.~Ma, H.~Xu, J.~Jiang, X.~Mei, and X.-P. Zhang, ``Ddcgan: A dual-discriminator
  conditional generative adversarial network for multi-resolution image
  fusion,'' \emph{{IEEE} TIP}, vol.~29, pp. 4980--4995, 2020.

\bibitem{RFN}
H.~Li, X.~Wu, and J.~Kittler, ``Rfn-nest: An end-to-end residual fusion network
  for infrared and visible images,'' \emph{Information Fusion}, vol.~73, pp.
  72--86, 2021.

\bibitem{MFEIF}
J.~Liu, X.~Fan, J.~Jiang, R.~Liu, and Z.~Luo, ``Learning a deep multi-scale
  feature ensemble and an edge-attention guidance for image fusion,''
  \emph{{IEEE} Transactions on Circuits and Systems for Video Technology},
  2021.

\bibitem{U2Fusion}
H.~Xu, J.~Ma, J.~Jiang, X.~Guo, and H.~Ling, ``U2fusion: {A} unified
  unsupervised image fusion network,'' \emph{{IEEE} Transactions of Pattern
  Analysis and Machine Intelligence}, vol.~44, no.~1, pp. 502--518, 2022.

\bibitem{flownet}
E.~Ilg, N.~Mayer, T.~Saikia, M.~Keuper, A.~Dosovitskiy, and T.~Brox, ``Flownet
  2.0: Evolution of optical flow estimation with deep networks,'' in
  \emph{CVPR}, 2017, pp. 1647--1655.

\bibitem{voxelmorph}
G.~Balakrishnan, A.~Zhao, M.~R. Sabuncu, J.~V. Guttag, and A.~V. Dalca, ``An
  unsupervised learning model for deformable medical image registration,'' in
  \emph{CVPR}, 2018, pp. 9252--9260.

\bibitem{dgcnet_19}
I.~Melekhov, A.~Tiulpin, T.~Sattler, M.~Pollefeys, E.~Rahtu, and J.~Kannala,
  ``Dgc-net: Dense geometric correspondence network,'' in \emph{WACV}, 2019,
  pp. 1034--1042.

\bibitem{glunet_20}
P.~Truong, M.~Danelljan, and R.~Timofte, ``Glu-net: Global-local universal
  network for dense flow and correspondences,'' in \emph{CVPR}, 2020, pp.
  6257--6267.

\bibitem{crossRAFT_22}
S.~Zhou, W.~Tan, and B.~Yan, ``Promoting single-modal optical flow network for
  diverse cross-modal flow estimation,'' in \emph{Thirty-Sixth {AAAI}
  Conference on Artificial Intelligence, {AAAI} 2022, Thirty-Fourth Conference
  on Innovative Applications of Artificial Intelligence, {IAAI} 2022, The
  Twelveth Symposium on Educational Advances in Artificial Intelligence, {EAAI}
  2022 Virtual Event, February 22 - March 1, 2022}, 2022, pp. 3562--3570.

\bibitem{SuperFusion_22}
L.~Tang, Y.~Deng, Y.~Ma, J.~Huang, and J.~Ma, ``Superfusion: A versatile image
  registration and fusion network with semantic awareness,'' \emph{{IEEE/CAA}
  Journal of Automatica Sinica}, vol.~9, no.~12, pp. 2121--2137, 2022.

\bibitem{NeMAR}
M.~Arar, Y.~Ginger, D.~Danon, A.~H. Bermano, and D.~Cohen{-}Or, ``Unsupervised
  multi-modal image registration via geometry preserving image-to-image
  translation,'' in \emph{CVPR}, 2020, pp. 13\,407--13\,416.

\bibitem{RFNet_22}
H.~Xu, J.~Ma, J.~Yuan, Z.~Le, and W.~Liu, ``Rfnet: Unsupervised network for
  mutually reinforcing multi-modal image registration and fusion,'' in
  \emph{CVPR}, 2022, pp. 19\,647--19\,656.

\bibitem{UMF}
D.~Wang, J.~Liu, X.~Fan, and R.~Liu, ``Unsupervised misaligned infrared and
  visible image fusion via cross-modality image generation and registration,''
  in \emph{International Joint Conference on Artificial Intelligence}, 2022,
  pp. 3508--3515.

\bibitem{DenseFuse}
H.~Li and X.~Wu, ``Densefuse: {A} fusion approach to infrared and visible
  images,'' \emph{{IEEE} Transactions on Image Processing}, vol.~28, no.~5, pp.
  2614--2623, 2019.

\bibitem{FusionDN20}
H.~Xu, J.~Ma, Z.~Le, J.~Jiang, and X.~Guo, ``Fusiondn: A unified densely
  connected network for image fusion,'' in \emph{AAAI}, 2020, pp.
  12\,484--12\,491.

\bibitem{DenseNet17}
G.~Huang, Z.~Liu, L.~van~der Maaten, and K.~Q. Weinberger, ``Densely connected
  convolutional networks,'' in \emph{CVPR}, 2017, pp. 2261--2269.

\bibitem{GANMcC}
J.~Ma, H.~Zhang, Z.~Shao, P.~Liang, and H.~Xu, ``Ganmcc: A generative
  adversarial network with multi-classification constraints for infrared and
  visible image fusion,'' \emph{{IEEE} TIM}, vol.~70, pp. 1--14, 2021.

\bibitem{SMoA_21}
J.~Liu, Y.~Wu, Z.~Huang, R.~Liu, and X.~Fan, ``Smoa: Searching a
  modality-oriented architecture for infrared and visible image fusion,''
  \emph{{IEEE} Signal Processing Letters}, vol.~28, pp. 1818--1822, 2021.

\bibitem{HAFA_21}
R.~Liu, Z.~Liu, J.~Liu, and X.~Fan, ``Searching a hierarchically aggregated
  fusion architecture for fast multi-modality image fusion,'' in \emph{{ACM}
  Multimedia Conference}, 2021, pp. 1600--1608.

\bibitem{LSLA_22}
J.~Liu, Y.~Wu, G.~Wu, R.~Liu, and X.~Fan, ``Learn to search a lightweight
  architecture for target-aware infrared and visible image fusion,''
  \emph{{IEEE} Signal Process. Lett.}, vol.~29, pp. 1614--1618, 2022.

\bibitem{SwinFusion}
J.~Ma, L.~Tang, F.~Fan, J.~Huang, X.~Mei, and Y.~Ma, ``Swinfusion: Cross-domain
  long-range learning for general image fusion via swin transformer,''
  \emph{{IEEE/CAA} Journal of Automatica Sinica}, vol.~9, no.~7, pp.
  1200--1217, 2022.

\bibitem{CGTF}
J.~Li, J.~Zhu, C.~Li, X.~Chen, and B.~Yang, ``{CGTF:} convolution-guided
  transformer for infrared and visible image fusion,'' \emph{{IEEE}
  Transactions on Instrumentation and Measurement}, vol.~71, pp. 1--14, 2022.

\bibitem{PPTFusion}
Y.~Fu, T.~Xu, X.~Wu, and J.~Kittler, ``{PPT} fusion: Pyramid patch transformer
  for a case study in image fusion,'' \emph{CoRR}, vol. abs/2107.13967, 2021.

\bibitem{SeAFusion}
L.~Tang, J.~Yuan, and J.~Ma, ``Image fusion in the loop of high-level vision
  tasks: {A} semantic-aware real-time infrared and visible image fusion
  network,'' \emph{Information Fusion}, vol.~82, pp. 28--42, 2022.

\bibitem{pwcnet_18}
D.~Sun, X.~Yang, M.~Liu, and J.~Kautz, ``Pwc-net: Cnns for optical flow using
  pyramid, warping, and cost volume,'' in \emph{CVPR}, 2018, pp. 8934--8943.

\bibitem{gmflow_22}
H.~Xu, J.~Zhang, J.~Cai, H.~Rezatofighi, and D.~Tao, ``Cvpr,'' 2022, pp.
  8111--8120.

\bibitem{Nemar_20}
M.~Arar, Y.~Ginger, D.~Danon, A.~H. Bermano, and D.~Cohen{-}Or, ``Unsupervised
  multi-modal image registration via geometry preserving image-to-image
  translation,'' in \emph{CVPR}, 2020, pp. 13\,407--13\,416.

\bibitem{scunet_22}
K.~Zhang, Y.~Li, J.~Liang, J.~Cao, Y.~Zhang, H.~Tang, R.~Timofte, and
  L.~Van~Gool, ``Practical blind denoising via swin-conv-unet and data
  synthesis,'' \emph{arXiv preprint}, 2022.

\bibitem{CA_18}
J.~Hu, L.~Shen, and G.~Sun, ``Squeeze-and-excitation networks,'' in
  \emph{CVPR}, 2018, pp. 7132--7141.

\bibitem{swinIR_21}
J.~Liang, J.~Cao, G.~Sun, K.~Zhang, L.~V. Gool, and R.~Timofte, ``Swinir: Image
  restoration using swin transformer,'' in \emph{ICCVW}, 2021, pp. 1833--1844.

\bibitem{SA_18}
S.~Woo, J.~Park, J.~Lee, and I.~S. Kweon, ``{CBAM:} convolutional block
  attention module,'' in \emph{ECCV}, V.~Ferrari, M.~Hebert, C.~Sminchisescu,
  and Y.~Weiss, Eds., vol. 11211, 2018, pp. 3--19.

\bibitem{SvSW}
S.~Ghosh, R.~G. Gavaskar, and K.~N. Chaudhury, ``Saliency guided image detail
  enhancement,'' in \emph{NCC}, 2019, pp. 1--6.

\bibitem{MSIFT}
M.~Brown and S.~S{\"u}sstrunk, ``Multi-spectral sift for scene category
  recognition,'' in \emph{CVPR}, 2011, pp. 177--184.

\bibitem{elastic}
P.~Y. Simard, D.~Steinkraus, and J.~C. Platt, ``Best practices for
  convolutional neural networks applied to visual document analysis,'' in
  \emph{ICDAR}, 2003, pp. 958--962.

\bibitem{MI_metric}
D.~Mahapatra, Z.~Ge, S.~Sedai, and R.~Chakravorty, ``Joint registration and
  segmentation of xray images using generative adversarial networks,'' in
  \emph{MICCAI}, 2018, pp. 73--80.

\bibitem{NCC_metric}
X.~Cao, J.~Yang, L.~Wang, Z.~Xue, Q.~Wang, and D.~Shen, ``Deep learning based
  inter-modality image registration supervised by intra-modality similarity,''
  in \emph{MLMI}, 2018, pp. 55--63.

\bibitem{vif}
Y.~Han, Y.~Cai, Y.~Cao, and X.~Xu, ``A new image fusion performance metric
  based on visual information fidelity,'' \emph{Information Fusion}, vol.~14,
  no.~2, pp. 127--135, 2013.

\bibitem{ssim}
Z.~Wang, A.~C. Bovik, H.~R. Sheikh, and E.~P. Simoncelli, ``Image quality
  assessment: from error visibility to structural similarity,'' \emph{{IEEE}
  Transactions on Image Processing}, vol.~13, no.~4, pp. 600--612, 2004.

\bibitem{DGCNet19}
I.~Melekhov, A.~Tiulpin, T.~Sattler, M.~Pollefeys, E.~Rahtu, and J.~Kannala,
  ``Dgc-net: Dense geometric correspondence network,'' in \emph{WACV}, 2019,
  pp. 1034--1042.

\bibitem{GLUNet20}
P.~Truong, M.~Danelljan, and R.~Timofte, ``Glu-net: Global-local universal
  network for dense flow and correspondences,'' in \emph{CVPR}, 2020, pp.
  6257--6267.

\bibitem{CrossRAFT22}
S.~Zhou, W.~Tan, and B.~Yan, ``Promoting single-modal optical flow network for
  diverse cross-modal flow estimation,'' in \emph{AAAI}, 2022, pp. 3562--3570.

\end{thebibliography}
\end{document}